\documentclass{article}

\usepackage{amsmath,amsfonts,bm}

\def\eqref#1{equation~\ref{#1}}

\def\1{\bm{1}}

\def\vb{{\bm{b}}}

\def\vk{{\bm{k}}}

\def\vs{{\bm{s}}}

\def\mM{{\bm{M}}}

\def\mS{{\bm{S}}}

\def\mW{{\bm{W}}}

\DeclareMathAlphabet{\mathsfit}{\encodingdefault}{\sfdefault}{m}{sl}
\SetMathAlphabet{\mathsfit}{bold}{\encodingdefault}{\sfdefault}{bx}{n}

\def\gB{{\mathcal{B}}}

\DeclareMathOperator*{\argmin}{arg\,min}

\usepackage{microtype}
\usepackage{graphicx}
\usepackage{subcaption}
\usepackage{booktabs} %
\usepackage{xcolor}
\usepackage{pifont}
\usepackage{array}

\usepackage{hyperref}

\usepackage[accepted]{icml2024}

\usepackage{amsmath}
\usepackage{amssymb}
\usepackage{mathtools}
\usepackage{amsthm}

\usepackage[capitalize,noabbrev]{cleveref}

\theoremstyle{plain}
\newtheorem{theorem}{Theorem}[section]

\theoremstyle{definition}

\theoremstyle{remark}

\usepackage[textsize=tiny]{todonotes}

\icmltitlerunning{Sparsest Models Elude Pruning}

\usepackage[ruled,vlined]{algorithm2e}
\usepackage{todonotes}
\usepackage[group-separator={,}]{siunitx}
\usepackage{enumitem}
\usepackage{pgfplots}
\usepackage{amssymb}
\usepackage{filecontents}

\pgfplotsset{compat=newest}

\newcommand{\PlotFigure}[5]{

\begin{tikzpicture}

\begin{axis}[
    xlabel={Nonzero Parameters},
    ylabel={Accuracy},
    ymin=50, ymax=100,
    ytick={50, 60, 70, 80, ...,100},
    ymajorgrids=true,
    grid style=dashed,
    grid=major,
    width=#5\linewidth,
    height=#5\linewidth,
    legend pos=outer north east,
    legend style={
        overlay,
        draw=none,
    },
]

\addplot[only marks, thick, color=\ColorComb, mark=*, mark options={scale=0.5}] table [col sep=comma, x=#2, y expr=100*\thisrow{#3}] {Appendix_Figures/Scatterplots/#1/comb_search.csv};

\addplot[only marks, thick, color=\ColorIMP, mark=*, mark options={scale=0.5}] table [col sep=comma, x=#2, y expr=100*\thisrow{#3}] {Appendix_Figures/Scatterplots/#1/imp.csv};

\addplot[only marks, thick, color=\ColorLTH, mark=*, mark options={scale=0.5}] table [col sep=comma, x=#2, y expr=100*\thisrow{#3}] {Appendix_Figures/Scatterplots/#1/imp_LTH.csv};

\addplot[only marks, thick, color=\ColorRIGL, mark=*, mark options={scale=0.5}] table [col sep=comma, x=#2, y expr=100*\thisrow{#3}] {Appendix_Figures/Scatterplots/#1/rigL.csv};

\addplot[only marks, thick, color=\ColorSynFlow, mark=*, mark options={scale=0.5}] table [col sep=comma, x=#2, y expr=100*\thisrow{#3}] {Appendix_Figures/Scatterplots/#1/synFlow.csv};

\addplot[only marks, thick, color=\ColorIterSNIP, mark=*, mark options={scale=0.5}] table [col sep=comma, x=#2, y expr=100*\thisrow{#3}] {Appendix_Figures/Scatterplots/#1/iter_snip.csv};

\addplot[only marks, thick, color=\ColorSNIP, mark=*, mark options={scale=0.5}] table [col sep=comma, x=#2, y expr=100*\thisrow{#3}] {Appendix_Figures/Scatterplots/#1/snip.csv};

\addplot[only marks, thick, color=\ColorProsPr, mark=*, mark options={scale=0.5}] table [col sep=comma, x=#2, y expr=100*\thisrow{#3}] {Appendix_Figures/Scatterplots/#1/prospr.csv};

\addplot[only marks, thick, color=\ColorGraSP, mark=*, mark options={scale=0.5}] table [col sep=comma, x=#2, y expr=100*\thisrow{#3}] {Appendix_Figures/Scatterplots/#1/grasp.csv};

\addplot[only marks, thick, color=\ColorFORCE, mark=*, mark options={scale=0.5}] table [col sep=comma, x=#2, y expr=100*\thisrow{#3}] {Appendix_Figures/Scatterplots/#1/FORCE.csv};

\def\argument{#4}
\def\true{true}
\ifx\argument\true
\addlegendentry{Comb.}
\addlegendentry{GMP}
\addlegendentry{LTH}
\addlegendentry{RigL}
\addlegendentry{SynFlow}
\addlegendentry{Iter SNIP}
\addlegendentry{SNIP}
\addlegendentry{ProsPr}
\addlegendentry{GraSP}
\addlegendentry{FORCE}
\else
\fi

\end{axis}
\end{tikzpicture}
}

\newcommand{\PlotTwoFigures}[5]{

\makebox[0.1\linewidth]{%
\begin{tikzpicture}

\begin{groupplot}[
    group style={
        group size=2 by 1,
        ylabels at=edge left,
        yticklabels at=edge left,
        horizontal sep=0.3cm,
    },
    ylabel={Accuracy},
    ymin=50, ymax=100,
    ytick={50, 55, ...,100},
    max space between ticks=15pt,
    ymajorgrids=true,
    grid style=dashed,
    grid=major,
    every tick label/.append style={font=\tiny},
    width=0.565\linewidth,
    height=0.565\linewidth,
    legend style={
        font=\scriptsize,
        at={(-0.32,1.08)},
        anchor=south west,
        legend columns=5,
        overlay,
        draw=none,
    },
]

\nextgroupplot[xlabel={Number of Nonzero Parameters}, xlabel style={font=\tiny}]
\PlotCommands{#1}{#3}{#4}{#5}

\nextgroupplot[xlabel={Number of Nonzero Parameters}, xlabel style={font=\tiny}]
\PlotCommands{#2}{#3}{#4}{false}

\end{groupplot}
\end{tikzpicture}
}
}

\newcommand{\PlotCommands}[4]{
\addplot[thick, color=\ColorComb, mark=*, mark options={scale=0.5}] table [col sep=comma, x=#2, y expr=100*\thisrow{#3}] {Figures/#1/comb_search.csv};

\addplot[thick, color=\ColorIMP, mark=*, mark options={scale=0.5}] table [col sep=comma, x=#2, y expr=100*\thisrow{#3}] {Figures/#1/imp.csv};

\addplot[thick, color=\ColorLTH, mark=*, mark options={scale=0.5}] table [col sep=comma, x=#2, y expr=100*\thisrow{#3}] {Figures/#1/imp_LTH.csv};

\addplot[thick, color=\ColorRIGL, mark=*, mark options={scale=0.5}] table [col sep=comma, x=#2, y expr=100*\thisrow{#3}] {Figures/#1/rigL.csv};

\addplot[thick, color=\ColorSynFlow, mark=*, mark options={scale=0.5}] table [col sep=comma, x=#2, y expr=100*\thisrow{#3}] {Figures/#1/synFlow.csv};

\addplot[thick, color=\ColorIterSNIP, mark=*, mark options={scale=0.5}] table [col sep=comma, x=#2, y expr=100*\thisrow{#3}] {Figures/#1/iter_snip.csv};

\addplot[thick, color=\ColorSNIP, mark=*, mark options={scale=0.5}] table [col sep=comma, x=#2, y expr=100*\thisrow{#3}] {Figures/#1/snip.csv};

\addplot[thick, color=\ColorProsPr, mark=*, mark options={scale=0.5}] table [col sep=comma, x=#2, y expr=100*\thisrow{#3}] {Figures/#1/prospr.csv};

\addplot[thick, color=\ColorGraSP, mark=*, mark options={scale=0.5}] table [col sep=comma, x=#2, y expr=100*\thisrow{#3}] {Figures/#1/grasp.csv};

\addplot[thick, color=\ColorFORCE, mark=*, mark options={scale=0.5}] table [col sep=comma, x=#2, y expr=100*\thisrow{#3}] {Figures/#1/FORCE.csv};

\def\argument{#4}
\def\true{true}
\ifx\argument\true
\addlegendentry{Comb.}
\addlegendentry{GMP}
\addlegendentry{LTH}
\addlegendentry{RigL}
\addlegendentry{SynFlow}
\addlegendentry{Iter SNIP}
\addlegendentry{SNIP}
\addlegendentry{ProsPr}
\addlegendentry{GraSP}
\addlegendentry{FORCE}
\else
\fi
}

\newcommand{\PlotTwoFiguresHighAcc}[5]{
\makebox[0.1\linewidth]{%
\begin{tikzpicture}

\begin{groupplot}[
    group style={
        group size=2 by 1,
        ylabels at=edge left,
        yticklabels at=edge left,
        horizontal sep=0.2cm,
    },
    ylabel={Accuracy},
    ymin=90, ymax=100,
    ytick={90, 91, ...,100},
    ymajorgrids=true,
    grid style=dashed,
    grid=major,
    width=0.59\linewidth,
    height=0.59\linewidth,
    legend style={
        font=\scriptsize,
        at={(-0.24,1.08)},
        anchor=south west,
        legend columns=5,
        overlay,
        draw=none,
    },
]

\nextgroupplot[xlabel={Number of Nonzero Parameters}, xlabel style={font=\tiny}]
\PlotCommands{#1}{#3}{#4}{#5}

\nextgroupplot[xlabel={Number of Nonzero Parameters}, xlabel style={font=\tiny}]
\PlotCommands{#2}{#3}{#4}{false}

\end{groupplot}
\end{tikzpicture}
}
}

\newcommand{\PlotTwoFiguresApp}[5]{

\begin{tikzpicture}

\begin{groupplot}[
    group style={
        group size=2 by 1,
        ylabels at=edge left,
        yticklabels at=edge left,
        horizontal sep=0.1cm,
    },
    ylabel={Accuracy},
    ymin=50, ymax=100,
    ytick={50, 55, ...,100},
    ymajorgrids=true,
    grid style=dashed,
    grid=major,
    width=0.55\linewidth,
    height=0.55\linewidth,
    legend style={
        at={(0.45,1.08)},
        anchor=south west,
        legend columns=5,
        draw=none,
    },
]

\nextgroupplot[xlabel={Nonzero Parameters}]
\PlotCommands{#1}{#3}{#4}{#5}

\nextgroupplot[xlabel={Nonzero Parameters}]
\PlotCommands{#2}{#3}{#4}{false}

\end{groupplot}
\end{tikzpicture}
}

\newcommand{\PlotSixFigures}[5]{

\begin{tikzpicture}

\begin{groupplot}[
    group style={
        group size=2 by 1,
        ylabels at=edge left,
        yticklabels at=edge left,
        horizontal sep=0.1cm,
    },
    ylabel={Accuracy},
    ymin=50, ymax=100,
    ytick={50, 55, ...,100},
    ymajorgrids=true,
    grid style=dashed,
    grid=major,
    width=0.402\linewidth,
    height=0.402\linewidth,
    legend style={
        at={(0.15,1.08)},
        anchor=south west,
        legend columns=5,
        overlay,
        draw=none,
    },
]

\nextgroupplot[xlabel={Nonzero Parameters}]
\PlotCommandsAppendix{#1}{#3}{#4}{#5}

\nextgroupplot[xlabel={Nonzero Parameters}]
\PlotCommandsAppendix{#2}{#3}{#4}{false}

\end{groupplot}
\end{tikzpicture}
}

\newcommand{\PlotCommandsAppendix}[4]{
\addplot[thick, color=\ColorComb, mark=*, mark options={scale=0.5}] table [col sep=comma, x=#2, y expr=100*\thisrow{#3}] {Appendix_Figures/#1/comb_search.csv};

\addplot[thick, color=\ColorIMP, mark=*, mark options={scale=0.5}] table [col sep=comma, x=#2, y expr=100*\thisrow{#3}] {Appendix_Figures/#1/imp.csv};

\addplot[thick, color=\ColorLTH, mark=*, mark options={scale=0.5}] table [col sep=comma, x=#2, y expr=100*\thisrow{#3}] {Appendix_Figures/#1/imp_LTH.csv};

\addplot[thick, color=\ColorRIGL, mark=*, mark options={scale=0.5}] table [col sep=comma, x=#2, y expr=100*\thisrow{#3}] {Appendix_Figures/#1/rigL.csv};

\addplot[thick, color=\ColorSynFlow, mark=*, mark options={scale=0.5}] table [col sep=comma, x=#2, y expr=100*\thisrow{#3}] {Appendix_Figures/#1/synFlow.csv};

\addplot[thick, color=\ColorIterSNIP, mark=*, mark options={scale=0.5}] table [col sep=comma, x=#2, y expr=100*\thisrow{#3}] {Appendix_Figures/#1/iter_snip.csv};

\addplot[thick, color=\ColorSNIP, mark=*, mark options={scale=0.5}] table [col sep=comma, x=#2, y expr=100*\thisrow{#3}] {Appendix_Figures/#1/snip.csv};

\addplot[thick, color=\ColorProsPr, mark=*, mark options={scale=0.5}] table [col sep=comma, x=#2, y expr=100*\thisrow{#3}] {Appendix_Figures/#1/prospr.csv};

\addplot[thick, color=\ColorGraSP, mark=*, mark options={scale=0.5}] table [col sep=comma, x=#2, y expr=100*\thisrow{#3}] {Appendix_Figures/#1/grasp.csv};

\addplot[thick, color=\ColorFORCE, mark=*, mark options={scale=0.5}] table [col sep=comma, x=#2, y expr=100*\thisrow{#3}] {Appendix_Figures/#1/FORCE.csv};

\def\argument{#4}
\def\true{true}
\ifx\argument\true
\addlegendentry{Comb.}
\addlegendentry{GMP}
\addlegendentry{LTH}
\addlegendentry{RigL}
\addlegendentry{SynFlow}
\addlegendentry{Iter SNIP}
\addlegendentry{SNIP}
\addlegendentry{ProsPr}
\addlegendentry{GraSP}
\addlegendentry{FORCE}
\else
\fi
}

\newcommand{\PlotBar}[0]{
\resizebox{0.95\linewidth}{!}{%
\begin{tikzpicture}
    \begin{axis}[
        /pgf/number format/1000 sep={},
        width=14cm, %
        height=7cm, %
        scale only axis,
        clip=false,
        separate axis lines,
        axis on top,
        xmin=0,
        xmax=11,
        xtick={1,..., 10}, %
        x tick style={draw=none},
        xticklabels={SynFlow,FORCE,SNIP,Iter SNIP, ProsPr, GraSP, RigL, GMP, Dense, LTH}, %
        ymin=1e8,
        ymax=4e9,
        ylabel={FLOPS},
        every axis plot/.append style={
          ybar,
          bar width=0.9cm, %
          fill,
          draw=none, %
        },
        axis line style={-}, %
        tick label style={font=\small}, %
        grid=none, %
        nodes near coords, %
        nodes near coords style={
            black,
            font=\tiny,
            /pgf/number format/fixed zerofill,
            /pgf/number format/precision=1, %
        }, %
      ]
      \addplot[\ColorSynFlow] coordinates {(1,187225649.222222)}; %
      \addplot[\ColorFORCE] coordinates{(2,190892643.555556)}; %
      \addplot[\ColorSNIP] coordinates{(4,193101952)}; %
      \addplot[\ColorIterSNIP] coordinates{(3,192558037.333333)}; %
      \addplot[\ColorProsPr] coordinates{(5,194999808)}; %
      \addplot[\ColorGraSP] coordinates{(6,202514219.111111)}; %
      \addplot[\ColorRIGL] coordinates{(7,209094513.777778)}; %
      \addplot[\ColorIMP] coordinates{(8,533424725.333333)}; %
      \addplot[black] coordinates{(9,1520064000.0)}; %
      \addplot[\ColorLTH] coordinates{(10,3423125333.33333)}; %
    \end{axis}
  \end{tikzpicture}
  }
}

\usepackage{tikz}
\usepackage{amsmath}
\usepackage{subcaption}
\usetikzlibrary{decorations.pathreplacing,decorations.pathmorphing}

\definecolor{pastelblue}{rgb}{0.69, 0.88, 0.9}
\definecolor{pastelred}{rgb}{1.0, 0.71, 0.76}
\definecolor{pastelgreen}{rgb}{0.59, 0.98, 0.59}

\definecolor{lighter}{rgb}{0.98, 0.73, 0.77}
\definecolor{darker}{rgb}{0.75, 0.22, 0.3}

\newcommand{\drawMatrixOne}[9]{%
    \pgfmathtruncatemacro\endRow{min(#7,#3)}
    \pgfmathtruncatemacro\endCol{min(#7,#4)}
    \foreach \i in {1,...,\endRow}{
        \foreach \j in {1,...,\endCol}{
            \fill[lighter] (#1+#5*\j-#5,#2-#5*\i) rectangle (#1+#5*\j,#2-#5*\i+#5);
        }
    }
    \pgfmathsetmacro\maxdim{max(#3,#4)*#5}
    \node[yshift=0cm] at (#1+#5*#4/2, #2-\maxdim-1.25*#5) {#6};

    \ifnum#7<#3+1
    \draw[decorate,decoration={brace,amplitude=5pt,raise=-0.21cm}]
        (#1-#5,#2-#5*#7) -- (#1-#5,#2) node[midway,xshift=-0.25cm] {#8};
    \fi
    \ifnum#7<#4+1
    \draw[decorate,decoration={brace,amplitude=5pt,raise=1pt}]
        (#1,#2) -- (#1+#5*#7,#2) node[midway,yshift=0.45cm] {#9};
    \fi
}

\newcommand{\drawMatrixTopTwo}[9]{%
    \pgfmathtruncatemacro\endRow{min(2,#3)}
    \pgfmathtruncatemacro\endCol{min(4,#4)}
    \foreach \i in {1,...,\endRow}{
        \foreach \j in {1,...,\endCol}{
            \fill[lighter] (#1+#5*\j-#5,#2-#5*\i) rectangle (#1+#5*\j,#2-#5*\i+#5);
        }
    }
    \pgfmathsetmacro\maxdim{max(#3,#4)*#5}
    \node[yshift=0cm] at (#1+#5*#4/2, #2-\maxdim-1.25*#5) {#6};

    \ifnum2<#3+1
    \draw[decorate,decoration={brace,amplitude=5pt,raise=-0.21cm}]
        (#1-#5,#2-#5*2) -- (#1-#5,#2) node[midway,xshift=-0.25cm] {#8};
    \fi
    \ifnum4<#4+1
    \draw[decorate,decoration={brace,amplitude=5pt,raise=1pt}]
        (#1,#2) -- (#1+#5*4,#2) node[midway,yshift=0.45cm] {#9};
    \fi
}

\newcommand{\drawMatrixTopThree}[9]{%
    \pgfmathtruncatemacro\endRow{min(4,#3)}
    \pgfmathtruncatemacro\endCol{min(4,#4)}
    \foreach \i in {1,...,\endRow}{
        \foreach \j in {1,...,\endCol}{
            \fill[lighter] (#1+#5*\j-#5,#2-#5*\i) rectangle (#1+#5*\j,#2-#5*\i+#5);
        }
    }
    \pgfmathsetmacro\maxdim{max(#3,#4)*#5}
    \node[yshift=0cm] at (#1+#5*#4/2, #2-\maxdim-1.25*#5) {#6};

    \ifnum4<#3+1
    \draw[decorate,decoration={brace,amplitude=5pt,raise=-0.21cm}]
        (#1-#5,#2-#5*4) -- (#1-#5,#2) node[midway,xshift=-0.25cm] {#8};
    \fi
    \ifnum4<#4+1
    \draw[decorate,decoration={brace,amplitude=5pt,raise=1pt}]
        (#1,#2) -- (#1+#5*4,#2) node[midway,yshift=0.45cm] {#9};
    \fi
}

\newcommand{\drawMatrixTopFour}[9]{%
    \pgfmathtruncatemacro\endRow{min(4,#3)}
    \pgfmathtruncatemacro\endCol{min(2,#4)}
    \foreach \i in {1,...,\endRow}{
        \foreach \j in {1,...,\endCol}{
            \fill[lighter] (#1+#5*\j-#5,#2-#5*\i) rectangle (#1+#5*\j,#2-#5*\i+#5);
        }
    }
    \pgfmathsetmacro\maxdim{max(#3,#4)*#5}
    \node[yshift=0cm] at (#1+#5*#4/2, #2-\maxdim-1.25*#5) {#6};

    \ifnum4<#3+1
    \draw[decorate,decoration={brace,amplitude=5pt,raise=-0.21cm}]
        (#1-#5,#2-#5*4) -- (#1-#5,#2) node[midway,xshift=-0.25cm] {#8};
    \fi
    \ifnum#7<#4+1
    \draw[decorate,decoration={brace,amplitude=5pt,raise=1pt}]
        (#1,#2) -- (#1+#5*#7,#2) node[midway,yshift=0.45cm] {#9};
    \fi
}

\newcommand{\drawMatrixTwo}[6]{%
    \edef\tempList{#6}
    \foreach \x/\y/\col in \tempList {
        \fill[\col] (#1+#5*\y-#5,#2-#5*\x) rectangle (#1+#5*\y,#2-#5*\x+#5);
    }
    \foreach \i in {1,...,#3} {
        \foreach \j in {1,...,#4} {
            \draw (#1+#5*\j-#5,#2-#5*\i) rectangle (#1+#5*\j,#2-#5*\i+#5);
        }
    }
}

\pgfkeys{
    /matrixdrawing/.cd,
    dim/.store in=\Dim,
    boxsize/.store in=\BoxSize,
    k/.store in=\K,
    myspace/.store in=\myspace,
    firstmatrixcolors/.store in=\FirstMatrixColors,
    secondmatrixcolors/.store in=\SecondMatrixColors,
    thirdmatrixcolors/.store in=\ThirdMatrixColors,
    fourthmatrixcolors/.store in=\FourthMatrixColors,
}

\newcommand{\myplot}[1]{%
    \pgfkeys{/matrixdrawing/.cd,#1}
    \begin{tikzpicture}[baseline=(current bounding box.north)]

        \drawMatrixOne{0}{0}{2}{\Dim}{\BoxSize}{$\mW^{[1]^{\top}}$}{\K}{}{$d^{[1]}$}
        \drawMatrixTwo{0}{0}{2}{\Dim}{\BoxSize}{\FirstMatrixColors}
        
        \drawMatrixOne{1*\Dim*\BoxSize+1*\myspace}{0}{\Dim}{\Dim}{\BoxSize}{$\mW^{[2]^{\top}}$}{\K}{$d^{[1]}$}{$d^{[2]}$}
        \drawMatrixTwo{1*\Dim*\BoxSize+1*\myspace}{0}{\Dim}{\Dim}{\BoxSize}{\SecondMatrixColors}
        
        \drawMatrixOne{2*\Dim*\BoxSize+2*\myspace}{0}{\Dim}{\Dim}{\BoxSize}{$\mW^{[3]^{\top}}$}{\K}{$d^{[2]}$}{$d^{[3]}$}
        \drawMatrixTwo{2*\Dim*\BoxSize+2*\myspace}{0}{\Dim}{\Dim}{\BoxSize}{\ThirdMatrixColors}

        \drawMatrixOne{3*\Dim*\BoxSize+3*\myspace}{0}{\Dim}{1}{\BoxSize}{$\mW^{[4]^{\top}}$}{\K}{$d^{[3]}$}{}
        \drawMatrixTwo{3*\Dim*\BoxSize+3*\myspace}{0}{\Dim}{1}{\BoxSize}{\FourthMatrixColors}
        
    \end{tikzpicture}
}

\newcommand{\myplottwo}[1]{%
    \pgfkeys{/matrixdrawing/.cd,#1}
    \begin{tikzpicture}[baseline=(current bounding box.north)]

        \drawMatrixOne{0}{0}{2}{\Dim}{\BoxSize}{$\mW^{[1]^{\top}}$}{2}{}{$d^{[1]}$}
        \drawMatrixTwo{0}{0}{2}{\Dim}{\BoxSize}{\FirstMatrixColors}
        
        \drawMatrixTopTwo{1*\Dim*\BoxSize+1*\myspace}{0}{\Dim}{\Dim}{\BoxSize}{$\mW^{[2]^{\top}}$}{\K}{$d^{[1]}$}{$d^{[2]}$}
        \drawMatrixTwo{1*\Dim*\BoxSize+1*\myspace}{0}{\Dim}{\Dim}{\BoxSize}{\SecondMatrixColors}
        
        \drawMatrixTopThree{2*\Dim*\BoxSize+2*\myspace}{0}{\Dim}{\Dim}{\BoxSize}{$\mW^{[3]^{\top}}$}{\K}{$d^{[2]}$}{$d^{[3]}$}
        \drawMatrixTwo{2*\Dim*\BoxSize+2*\myspace}{0}{\Dim}{\Dim}{\BoxSize}{\ThirdMatrixColors}

        \drawMatrixTopFour{3*\Dim*\BoxSize+3*\myspace}{0}{\Dim}{1}{\BoxSize}{$\mW^{[4]^{\top}}$}{\K}{$d^{[3]}$}{}
        \drawMatrixTwo{3*\Dim*\BoxSize+3*\myspace}{0}{\Dim}{1}{\BoxSize}{\FourthMatrixColors}
        
    \end{tikzpicture}
}

\pgfplotsset{every axis/.append style={font=\footnotesize}}

\usetikzlibrary{pgfplots.groupplots}

\definecolor{darkblue}{RGB}{31,119,180}
\definecolor{lightblue}{RGB}{174,199,232}
\definecolor{darkorange}{RGB}{255,127,14}
\definecolor{lightorange}{RGB}{255,187,120}
\definecolor{darkgreen}{RGB}{44,160,44}
\definecolor{lightgreen}{RGB}{152,223,138}
\definecolor{darkred}{RGB}{214,39,40}
\definecolor{lightred}{RGB}{255,152,150}
\definecolor{darkpurple}{RGB}{148,103,189}
\definecolor{lightpurple}{RGB}{197,176,213}
\definecolor{darkbrown}{RGB}{140,86,75}
\definecolor{lightbrown}{RGB}{196,156,148}
\definecolor{darkpink}{RGB}{227,119,194}
\definecolor{lightpink}{RGB}{247,182,210}
\definecolor{darkgray}{RGB}{127,127,127}
\definecolor{lightgray}{RGB}{199,199,199}
\definecolor{darkolive}{RGB}{188,189,34}
\definecolor{lightolive}{RGB}{219,219,141}
\definecolor{darkcyan}{RGB}{23,190,207}
\definecolor{lightcyan}{RGB}{158,218,229}

\newcommand{\ColorComb}{darkblue}
\newcommand{\ColorIMP}{darkorange}
\newcommand{\ColorLTH}{darkgreen}
\newcommand{\ColorRIGL}{darkred}
\newcommand{\ColorSynFlow}{darkpurple}
\newcommand{\ColorIterSNIP}{darkbrown}
\newcommand{\ColorSNIP}{lightbrown}
\newcommand{\ColorProsPr}{darkpink}
\newcommand{\ColorGraSP}{darkgray}
\newcommand{\ColorFORCE}{lightgray} %

\SetKwProg{Fn}{Function}{\string:}{}
\SetKwFunction{CombinatorialSearch}{\textsc{CombinatorialSearch}}

\begin{document}

\twocolumn[
\icmltitle{Sparsest Models Elude Pruning: An Exposé of Pruning’s Current Capabilities}

\icmlsetsymbol{equal}{*}

\begin{icmlauthorlist}
\icmlauthor{Stephen Zhang}{yyy}
\icmlauthor{Vardan Papyan}{yyy}
\end{icmlauthorlist}

\icmlaffiliation{yyy}{Department of Mathematics, University of Toronto, Toronto, Canada}

\icmlcorrespondingauthor{Stephen Zhang}{stephenn.zhang@mail.utoronto.ca}

\icmlkeywords{Machine Learning, ICML}

\vskip 0.3in
]

\printAffiliationsAndNotice{}  %

\begin{abstract}
Pruning has emerged as a promising approach for compressing large-scale models, yet its effectiveness in recovering the sparsest of models has not yet been explored. We conducted an extensive series of \num{485838} experiments, applying a range of state-of-the-art pruning algorithms to a synthetic dataset we created, named the Cubist Spiral. Our findings reveal a significant gap in performance compared to ideal sparse networks, which we identified through a novel combinatorial search algorithm. We attribute this performance gap to current pruning algorithms' poor behaviour under overparameterization, their tendency to induce disconnected paths throughout the network, and their propensity to get stuck at suboptimal solutions, even when given the optimal width and initialization. This gap is concerning, given the simplicity of the network architectures and datasets used in our study. We hope that our research encourages further investigation into new pruning techniques that strive for true network sparsity.
\end{abstract}

\section{Introduction}
The burgeoning complexity of state-of-the-art deep learning models has made their training and deployment prohibitively expensive. To counteract this increasing demand for resources, model compression has become increasingly important in optimizing the computational efficiency of these networks. Among these techniques, a popular and proven option is network pruning which induces sparsity in the model parameters \citep{heofler_prune_survey}. 

Whilst pruning is effective, it is difficult to assess how close current pruning algorithms are to obtaining the sparsest of models due to the complexity of the datasets and models used. For the same reason, analyzing and interpreting pruning's effects on trained models has been challenging which has allowed for potential shortcomings to go unnoticed.

 \begin{figure}[t]
    \includegraphics[width=\linewidth, trim=80 320 100 60, clip]{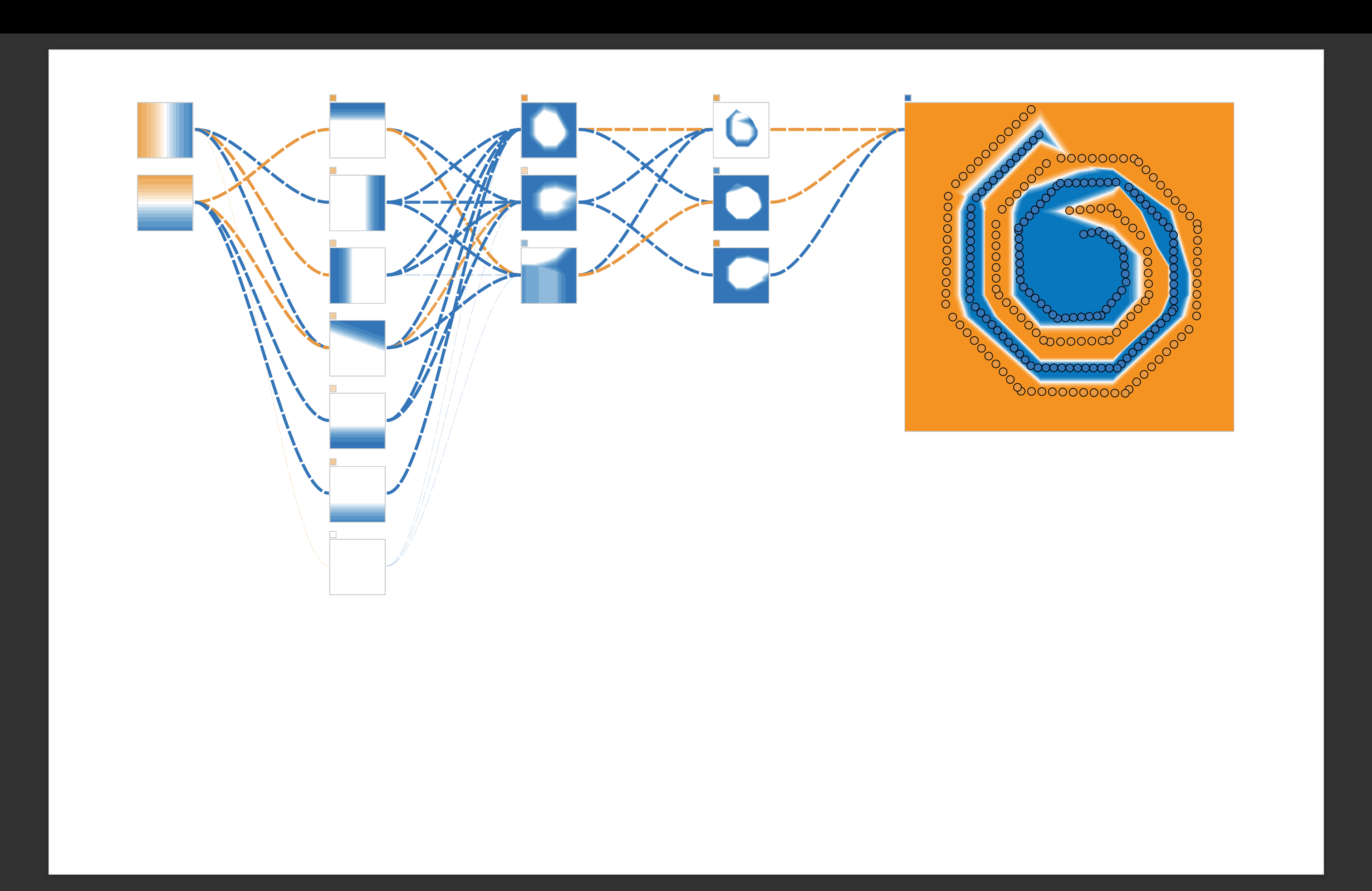}
    \caption{\textbf{Sparse Model Visualization.} Visualization of a sparse model, discovered through our combinatorial search algorithm, trained on the Cubist Spiral dataset. The first two squares on the left denote the input variables, while the final, larger square depicts the output from the classifier. The intermediate squares reveal post-activation states which are connected by edges, corresponding to entries in weight matrices. At the top of each square, there is a tiny square that is colored according to the bias of the corresponding neuron. Blue is used to represent a positive value, orange a negative value, and white -- a value of zero.}
    \label{fig:example_viz}
\end{figure}

This paper aims to scrutinize how closely various pruning algorithms approach the ideal, sparsest network, defined as the model with the fewest nonzero parameters that can achieve a specific target accuracy, and reveal the true efficacy of current pruning techniques.

\subsection{Method Overview}
To achieve our goals, we engineer the following tools that will be the basis for our analysis:
    \begin{description}
    \item[Cubist Spiral] A synthetic dataset named the \textit{Cubist Spiral}, depicted in Figure \ref{fig:cubist-spiral}. The simplicity inherent in the dataset leads to interpretable sparse models that are amenable to visualization and analysis.
    \item[Combinatorial Search] A novel combinatorial search algorithm that searches across model sparsity masks for an optimal and maximally sparse model. Diverging from existing naive benchmarks such as random pruning, where the pruned weights are selected randomly, our algorithm leverages structured sparsity to perform an efficient exploration across sparsity masks for the model.
     \item[Sparse Model Visualization] A visualization tool, similar to TensorFlow Playground \citep{smilkov2017direct}, designed for inspecting sparse models by graphically representing their non-zero subnetwork. An example of a model visualization is shown in Figure \ref{fig:example_viz}.
    \end{description}
\subsection{Contributions}
Through an empirical study (code available on \href{https://github.com/stephenqz/sparsest/}{GitHub}), we uncover the following deficiencies: %
\begin{description}
    \item[Algorithms Fail to Get Sparsest Model] There exists a disparity between the achievable outcomes and the current capabilities of pruning techniques in terms of recovering a sparsest model.  
    \item[Overparameterization Impedes Pruning] Unstructured pruning techniques are unable to adequately perform structured pruning resulting in a deterioration of their performance under overparameterization. 
    \item[Pruning Fails Under Optimal Conditions] Pruning is unable to recover the sparsest sparsity masks for the model even when provided with the optimal width and initialization.
\end{description}
    Through our visualization, we show:
    \begin{description}
    \item[Disconnected Paths] Pruning algorithms are unable to correctly align the parameters of consecutive layers resulting in disconnected paths. This leads to an inflated number of nonzero parameters that are not contributing to the expressivity of the network. 
    \item[Pruning Algorithms Foregoes Sparsity] Pruned networks can be further pruned after training without harming model performance. 
\end{description}

\section{Background: Pruning Algorithms}
Pruning algorithms are commonly classified into two main categories: structured and unstructured. In unstructured pruning, individual weights are pruned, whereas structured pruning operates at a higher level by pruning entire filters or channels \citep{Wen2016structure, li2017pruning, Luo_2017_ICCV}.

Beyond structured and unstructured, pruning algorithms can also be classified into the following three categories based on their pruning strategies.

\begin{figure}[t]
    \centering
    \begin{subfigure}[t]{0.48\linewidth}
        \centering
        \includegraphics[width=0.7\linewidth, trim=2cm 2cm 2cm 2cm, clip=true]{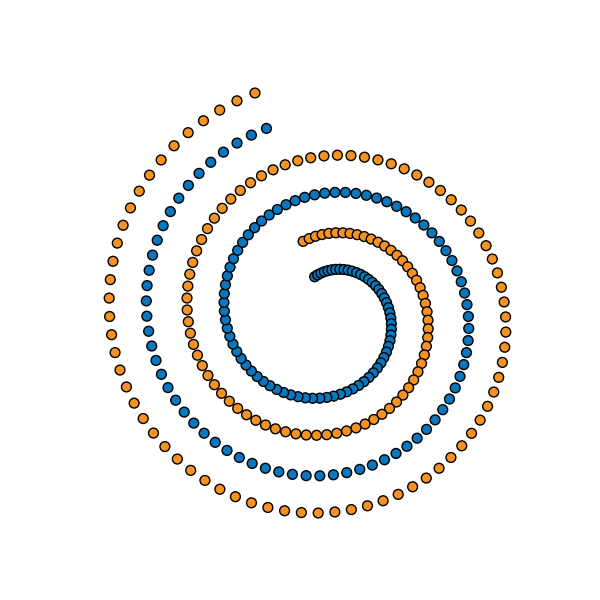}
        \caption{Spiral dataset}
        \label{fig:spiral}
    \end{subfigure}
    \hfill
    \begin{subfigure}[t]{0.48\linewidth}
        \centering
        \includegraphics[width=0.7\linewidth, trim=2cm 2cm 2cm 2cm, clip=true]{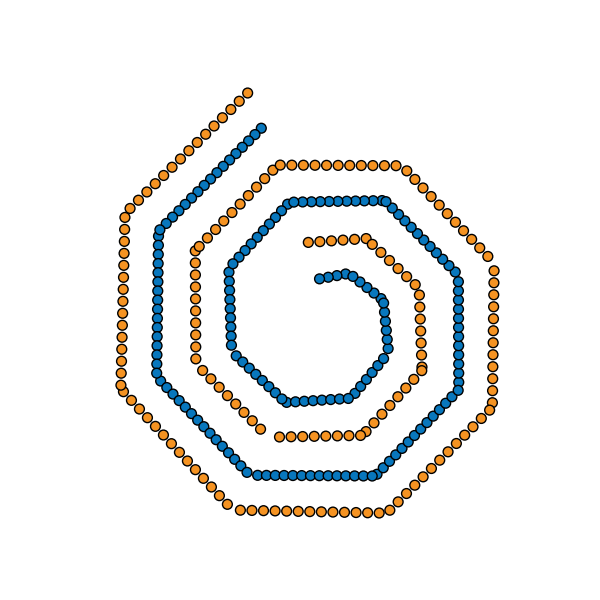}
        \caption{Cubist Spiral dataset}
        \label{fig:cubist-spiral}
    \end{subfigure}
    \caption{Comparative view of spiral datasets.}
    \label{fig:spiral-datasets}
    \end{figure}

\subsection{Dense to Sparse}
This category encompasses the methods of Optimal Brain Damage by \citet{lecun1989optimal} and its successor, Optimal Brain Surgeon by \citet{hassibi1992second,hassibi1993optimal}, which are seminal works not only in dense-to-sparse pruning but in pruning in general. More recently, magnitude-based pruning approaches have proven to be extremely effective \citep{han2015learning}, leading to state-of-the-art methods being developed such as Gradual Magnitude Pruning (GMP) by \citet{zhuGMP} and the Lottery Ticket Hypothesis (LTH) by \citet{frankle2018lottery}. These techniques typically start with a dense network configuration and implement pruning either progressively during the training process or upon its completion. While they offer resource savings during the inference stage, they do not reduce resource utilization during the training phase.

\subsection{Pruning at Initialization}
Representative methods in this category include Gradient Signal Preservation (GraSP) by \citet{wang2019picking}, Prospect Pruning (ProsPr) by \citet{alizadeh2021prospect}, Single-shot Network Pruning (SNIP) by \citet{lee2019snip}, Iterative Synaptic Flow Pruning (SynFlow) by \citet{tanaka2020pruning} and Iter SNIP and FORCE by \citet{jorge2021progressive}. In contrast to the previous category, these algorithms involve pruning neural networks at the initialization stage, followed by training the already-pruned models. This approach is beneficial as it conserves resources both during the training and inference phases, assuming the initial pruning overhead is negligible. 

\subsection{Sparse to Sparse}
Sparse Evolutionary Training (SET) by \citet{mocanu_scalable_2018} was the pioneer algorithm in this category. Subsequently, several other algorithms have been introduced, such as Deep-R by \citet{bellec2018deep}, Sparse Networks From Scratch (SNFS) by \citet{dettmers2019snfs}, and Dynamic Sparse Reparameterization (DSR) by \citet{pmlr-v97-mostafa19a}. The Rigged Lottery (RigL) by \citet{evci2020rigging} has emerged as a state-of-the-art method in this group. Distinguishing itself from other categories, this approach initiates with a sparsely connected neural network and maintains the total number of parameters while dynamically altering the nonzero connections throughout training. 

\begin{figure*}[t!]
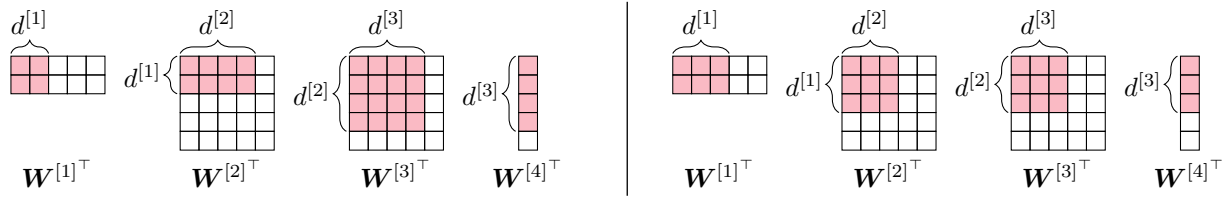

    \centering
    \begin{subfigure}[t]{0.48\textwidth}
        \centering
        \myplottwo{
            dim=5,
            boxsize=0.25,
            k=3,
            myspace=1,
            firstmatrixcolors=,
            secondmatrixcolors=,
            thirdmatrixcolors=,
            fourthmatrixcolors=,
        }
    \end{subfigure}
    \hspace{0.01\textwidth} %
    \vrule width 0.5pt %
    \hspace{0.01\textwidth} %
    \begin{subfigure}[t]{0.48\textwidth}
        \centering
        \myplot{
            dim=5,
            boxsize=0.25,
            k=3,
            myspace=1,
            firstmatrixcolors=,
            secondmatrixcolors=,
            thirdmatrixcolors=,
            fourthmatrixcolors=,
        }
    \end{subfigure}
    \caption{\textbf{First Phase.} Two structured sparsity masks that would be tested by the first phase of the combinatorial search. It is always the first $d^{[\ell-1]}$ columns and $d^{[\ell]}$ rows that are nonzero inside the masks, denoted by the light red squares. If both sets of masks reach the target accuracy, the set of masks on the right will be utilized by the second phase as it contains fewer nonzeros.}
    \label{fig:schematicPhaseOne}
\end{figure*}

\section{Methodology}
\subsection{Network}
The objective of this study is to evaluate the effectiveness of pruning algorithms in identifying the sparsest possible network. Finding it requires a combinatorial search which is only practical for smaller network architectures, due to its complexity. 

We therefore train four-layer Multilayer Perceptrons (MLPs) with ReLU activation functions, which take as input two coordinates and predicts a class label. 

All combinatorial search experiments are done on MLPs of width 16. The pruning algorithms, on the other hand, are run on MLPs of varying widths:
\[
\{3, 4, 5, 6, 7, 8, 16, 32, 64, 128, 256\},
\]
to examine the impact of overparameterization on their efficacy.

\subsection{Dataset}
The simplicity of the architecture calls for a simple dataset as well. We opt for the classical synthetic spiral dataset, notable for its non-linear separability. To better suit sparse modeling techniques, we have adapted the spiral by straightening its naturally curved edges. This modification gives rise to what we call the \textit{Cubist Spiral} dataset, a nod to the Cubism art movement that emphasized the use of minimal geometric shapes when depicting objects of interest. The classical spiral and its Cubist counterpart are juxtaposed in Figure \ref{fig:spiral-datasets}. 

We pick $50,000$ points spaced evenly along the spiral divided equally between the two classes. This deliberate choice of a large training set stems from our desire to separate any issues related to generalization when evaluating the efficacy of pruning algorithms. 

\subsection{Combinatorial Search}
\SetKwFunction{EligibleMasks}{\textsc{EligibleMasks}}
The combinatorial search is encapsulated in a function which obtains as input the width of the network $D$ and a desired target accuracy $\rho$ and returns a list of model sparsity masks for the MLP. The function involves two phases.

\paragraph{First Phase: Structured Sparsity}
The first phase performs a grid search over the number of neurons in each layer that span over the set $\{1,2, ..., D\}$. We denote the number of neurons in layer $\ell$ as $d^{[\ell]}$ with $d^{[0]}=2$ and $d^{[4]} = 1$. For each neuron configuration, a four-layer MLP is randomly initialized and masked such that only the first $d^{[\ell-1]}$ columns and $d^{[\ell]}$ rows in layer $\mW^{[\ell]}$ are nonzero. The MLP is then trained and a final accuracy is computed. Given the results from all the trainings, the configuration that achieves the desired target accuracy with the fewest nonzeros is selected. A schematic showing how phase one operates is displayed above in Figure \ref{fig:schematicPhaseOne}. 
\begin{figure*}[t!]
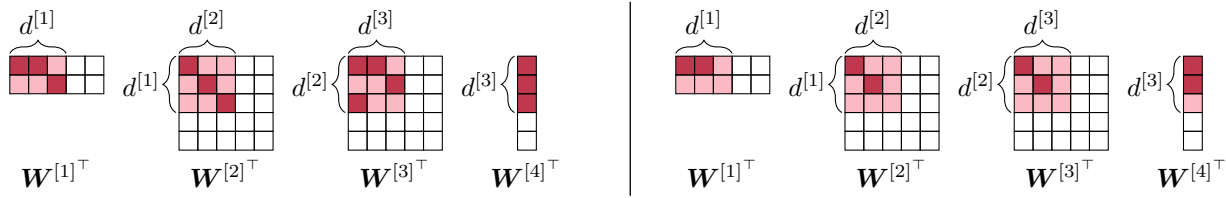

    \centering
    \begin{subfigure}[b]{0.48\textwidth}
        \centering
        \myplot{
            dim=5,
            boxsize=0.25,
            k=3,
            myspace=1,
            firstmatrixcolors={1/1/darker,1/2/darker,2/3/darker},
            secondmatrixcolors={1/1/darker,2/2/darker,3/3/darker},
            thirdmatrixcolors={1/1/darker,2/3/darker,3/1/darker,1/2/darker},
            fourthmatrixcolors={1/1/darker,2/1/darker,3/1/darker},
        }
    \end{subfigure}
    \hspace{0.01\textwidth} %
    \vrule width 0.5pt %
    \hspace{0.01\textwidth} %
    \begin{subfigure}[b]{0.48\textwidth}
        \centering
        \myplot{
            dim=5,
            boxsize=0.25,
            k=3,
            myspace=1,
            firstmatrixcolors={1/1/darker,1/2/darker},
            secondmatrixcolors={1/1/darker,2/2/darker},
            thirdmatrixcolors={1/1/darker,2/2/darker},
            fourthmatrixcolors={1/1/darker,2/1/darker},
        }
    \end{subfigure}
    \caption{\textbf{Second Phase.} A schematic illustration of the second phase of the combinatorial search. \textbf{Left:} A set of unstructured sparsity masks that would be tested in the second phase, generated by utilizing the minimal structured sparsity masks found in the first phase. The dark red squares denote the nonzero entries in the unstructured sparsity masks and the light red squares denote the nonzero entries of the minimal structured sparsity masks found in the first phase. \textbf{Right:} An ineligible mask containing rows and columns without at least one nonzero element, which does not fully utilize the minimal number of neurons identified in the first phase.}
    \label{fig:schematicPhaseTwo}
\end{figure*}
\paragraph{Second Phase: Unstructured Sparsity}
The second phase iterates over a list of unstructured sparsity masks for each weight matrix.\footnote{The combinatorial search iterates only over the masks of the weight matrices. As for the biases, we assign a value of zero to the $i$-th bias entry if and only if the $i$-th row of the weight matrix in that layer is zero in the mask.} For weight $\mW^{[\ell]}$, this list is generated by the function \(\EligibleMasks(d^{[\ell-1]}, d^{[\ell]})\) where $d^{[\ell-1]}$ and $d^{[\ell]}$ are determined based on the optimal configuration established in the first phase.

To ensure that the combinatorial search is done efficiently, $\EligibleMasks(d^{[\ell]}, d^{[\ell-1]})$ ensures that each mask for the weight is confined to the rows and columns essential for fulfilling the neuron configuration. Furthermore, each required row and column contains at least one nonzero element. This assumption is grounded in the notion that the neuron configuration, as determined in the first phase of the search, is inherently minimal.

$\EligibleMasks(d^{[\ell]}, d^{[\ell-1]})$  further optimizes the combinatorial search by eliminating masks that are functionally identical but differ merely by permutations of channels. The symmetry is broken by selecting from all row permutations the specific arrangement that results in a sequentially decreasing count of nonzero elements. If two rows have an equal count of nonzeros, the algorithm converts the binary vector representations of these masks into their decimal equivalents and arranges them in descending order. A schematic for phase two is depicted below in Figure \ref{fig:schematicPhaseTwo}.

This process provides a list of sparsity masks for each weight matrix which the combinatorial search then combines to form a list of masks for the model. These masks are then applied to models that are randomly initialized prior to any training. The algorithm for the combinatorial search is detailed in Algorithm \ref{alg:combinatorial} in the Appendix.

\subsection{Selection of Pruning Algorithms}
We benchmark the following pruning algorithms: GMP, LTH, GraSP, SNIP, SynFlow, Iter SNIP, FORCE, ProsPr, and RigL. We depict in Table \ref{table:prune_type} below each technique's categorization along with the FLOPS required to prune and train an MLP of width 16. We prune multiple models with different budgets of nonzeros for the weights. The budgets are specifically chosen to be centered around the range where the combinatorial search can reconstruct the spiral. Further details are provided in Appendix \ref{app:prune_hyper} and \ref{app:flops}. 

\begin{table}[ht]
    \centering
    \resizebox{\linewidth}{!}{%
    \begin{tabular}{@{}lcccccc@{}}
      \toprule
      \textbf{Pruning Techniques} & \multicolumn{2}{c}{\textbf{Pruning at Initialization}} & \textbf{Dynamic Sparse Training} & \textbf{Dense Training Required} & \textbf{FLOPS Required} \\ \cmidrule(lr){2-3}
                                  & \textbf{One Shot}  & \textbf{Iterative}                                &                                                  &                               &                        \\ \midrule
      LTH                         &                    &                                                   &                                                  & \centering \ding{51}           & $3.4\cdot 10^9$                   \\
      Dense Training                       &                    &                                                   &                                                  & \centering \ding{51}           & $1.5 \cdot 10^9$                   \\
      GMP                         &                    &                                                   &                                                  & \centering \ding{51}           & $5.3 \cdot 10^8$                   \\
      RigL                        &                    &                                                   & \centering \ding{51}                              &                               & $2.1 \cdot 10^8$                \\
      GraSP                       & \centering \ding{51} &                                                   &                                                  &                               & $2.0\cdot 10^8$                    \\
      ProsPr                      & \centering \ding{51} &                                                   &                                                  &                               & $1.9\cdot 10^8$                    \\
      SNIP                        & \centering \ding{51} &                                                   &                                                  &                               & $1.9\cdot 10^8$                    \\
      Iter SNIP                   &                    & \centering \ding{51}                               &                                                  &                               & $1.9\cdot 10^8$                    \\
      FORCE                       &                    & \centering \ding{51}                               &                                                  &                               & $1.9\cdot 10^8$                    \\
      SynFlow                     &                    & \centering \ding{51}                               &                                                  &                               & $1.9\cdot 10^8$                \\\bottomrule
    \end{tabular}%
    }
    \caption{Table depicting the categorization and FLOPS required for each pruning technique that was tested in our experiments.}
    \label{table:prune_type}
  \end{table}

The aforementioned pruning techniques do not include bias parameters in the pruning process. To ensure that the comparison to the combinatorial search is fair, entries of the bias are masked based on whether the corresponding column in the succeeding weight matrix is fully pruned, i.e., $b_i^{[l]}$ is masked if and only if $\mW^{[l+1]}_{:\;, \;i} = 0$. The bias corresponding to the classifier layer always remains fully dense.

\subsection{Initialization Experiments}\label{sec:init_exper}
The combinatorial search trains models on a large number of model masks, where each trained model starts at a different initialization of the parameters. One might speculate that the success of the combinatorial search could be tied to the initialization rather than the actual mask of the parameters.

To study the effect of the parameter initialization on the success of the pruning algorithms, we pick the best initialization from the combinatorial search -- the one that led to the sparsest model for a given target accuracy, $\rho$ -- and use it to initialize the pruning experiments. If a good initialization is all that is needed for successful pruning, then the pruning algorithms should succeed and be able to match the combinatorial search.

We equalize the comparison with the combinatorial search by running another round of the combinatorial search, but this time using the most successful initialization from the first combinatorial search to account for giving the pruned models the initialization. This also serves as a sanity check to verify whether the initialization is advantageous compared to a typical random initialization.

\subsection{Optimization}
We train the model parameters for 50 epochs using stochastic gradient descent (SGD) with momentum \num{0.9} and a batch size of \num{128}. Parameters outside of the determined model mask are constrained to be zero. A weight decay is applied for all experiments and set to $5\mathrm{e}{-4}$. For the pruning experiments, learning rates $\{0.05, 0.1, 0.2\}$ are used while for the combinatorial search, only $\{0.05, 0.1\}$ are used. We also utilize three learning rate schedulers: constant learning rate, a cosine annealing scheduler, and a decay of $0.1$ applied at epochs $15$ and $30$. 

\section{Combinatorial Search Results}
\subsection{Phase One of the Combinatorial Search}
 Preliminary experiments, which involve running the first phase of the combinatorial search with varying target sparsities, reveal three categories of model performance:
\begin{enumerate}
\item \textbf{Below $95\%$ Accuracy:} Models in this group were unable to even approximately reconstruct the spiral.
\item \textbf{Between $95\%$ and $99.5\%$ Accuracy:} Models within this range partially reconstructed the spiral as a sparse combination of polygons. However, they fell short of complete accuracy due to misclassification of certain minor segments of the spiral.
\item \textbf{Above $99.5\%$ Accuracy:} Models surpassing 99.5\% accuracy demonstrated essentially perfect reconstruction of the spiral.
\end{enumerate}

\begin{figure}[h]
        \centering
        \includegraphics[width=0.9\linewidth]{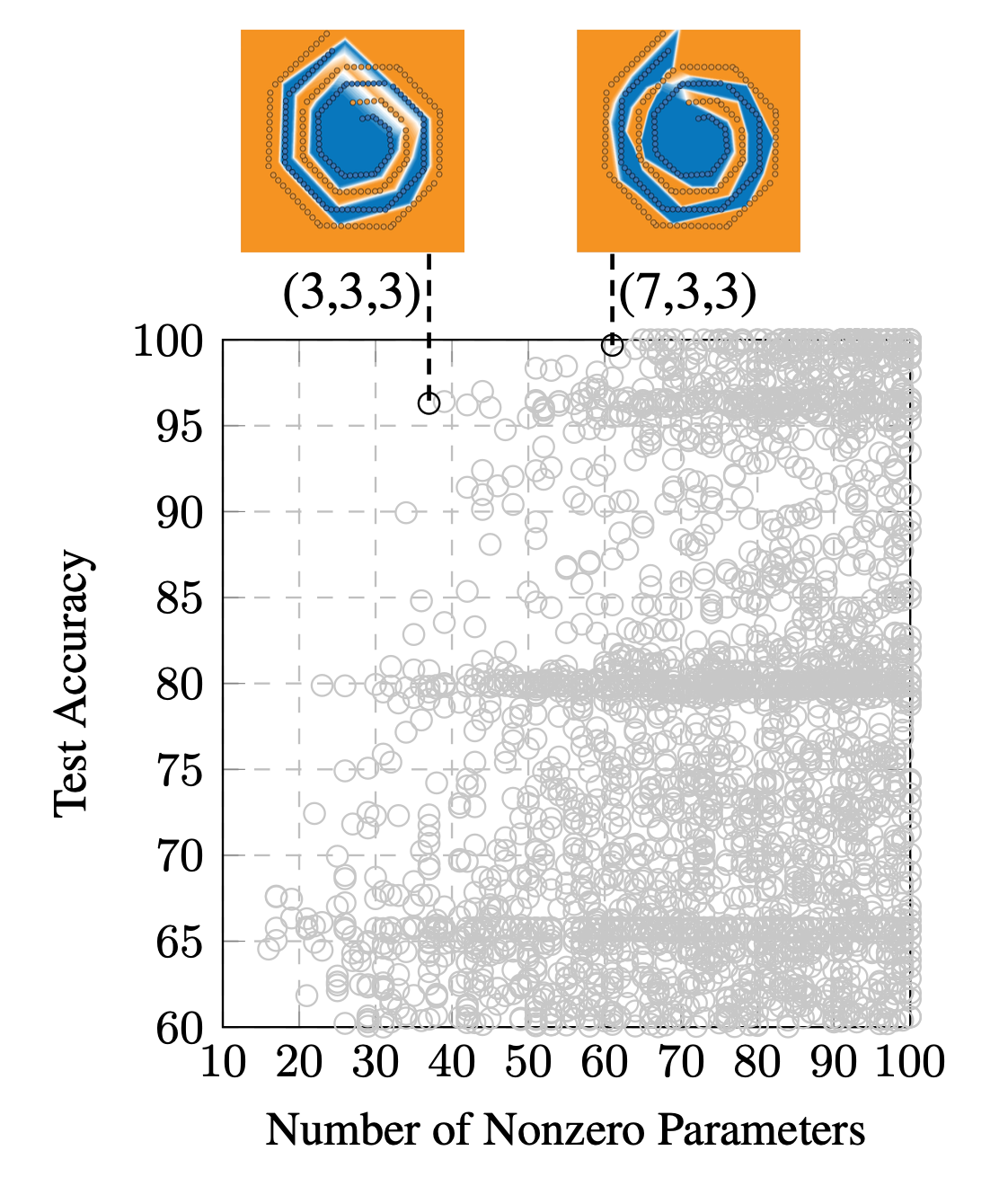}
        \caption{\textbf{Phase One of Combinatorial Search.} Scatter plot with each point corresponding to a different model that was trained with a different structured mask. Two models are highlighted -- the sparsest achieving above 95\% accuracy (where the number of neurons in each layer is 3,3,3) and the sparsest achieving above 99.5\% accuracy (where the number of neurons in each layer is 7,3,3) -- accompanied by their corresponding reconstructions of the spiral.}
        \label{fig:prelimWithSpiral}
\end{figure}

Based on this categorization, we rerun the combinatorial search twice; once with the target accuracy of $\rho_1= 95\%$ and a second time with $\rho_2=99.5\%$.  Figure \ref{fig:prelimWithSpiral} below contains a scatter plot of all the models trained in the first phase of the search.

\subsection{Phase Two of the Combinatorial Search}
Phase two of the algorithm provides a total of \num{25992} model masks to try for the 95\% target accuracy and \num{266004066} model masks for the 99.5\% target accuracy. Due to computational limits, we check only a subset of size \num{63208} of the possible model masks for the 99.5\% target accuracy. Details on the subset are given in Appendix \ref{app:995_subset}

For $\rho_1=95\%$, the combinatorial search found the model presented in Figure \ref{fig:bench_333} below. We refer to this benchmark model as \textsc{Bench-95}. 

\begin{figure}[h]
        \centering
        \includegraphics[width=\linewidth, trim=80 500 100 60, clip]{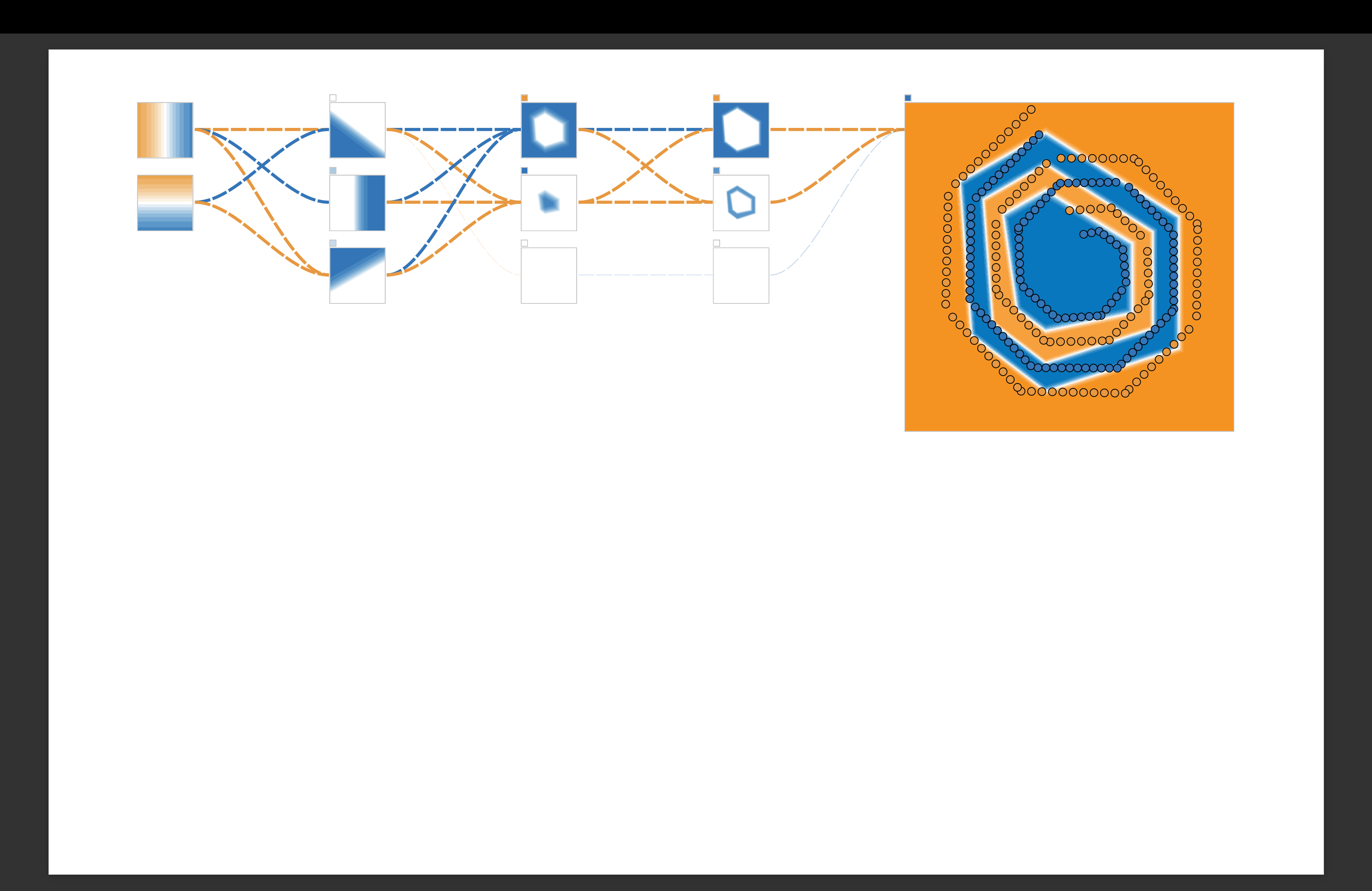}
        \caption{The minimal model found through the second phase of the combinatorial search that achieved over $95\%$ accuracy. The model has $30$ nonzero parameters and an accuracy of $96.04\%$.}
        \label{fig:bench_333}
\end{figure}

For $\rho_2 = 99.5\%$, the combinatorial search  found the model presented in Figure \ref{fig:bench_733} below. We refer to this benchmark model as \textsc{Bench-995}. 

\begin{figure}[h]
        \centering
        \includegraphics[width=\linewidth, trim=80 320 100 60, clip]{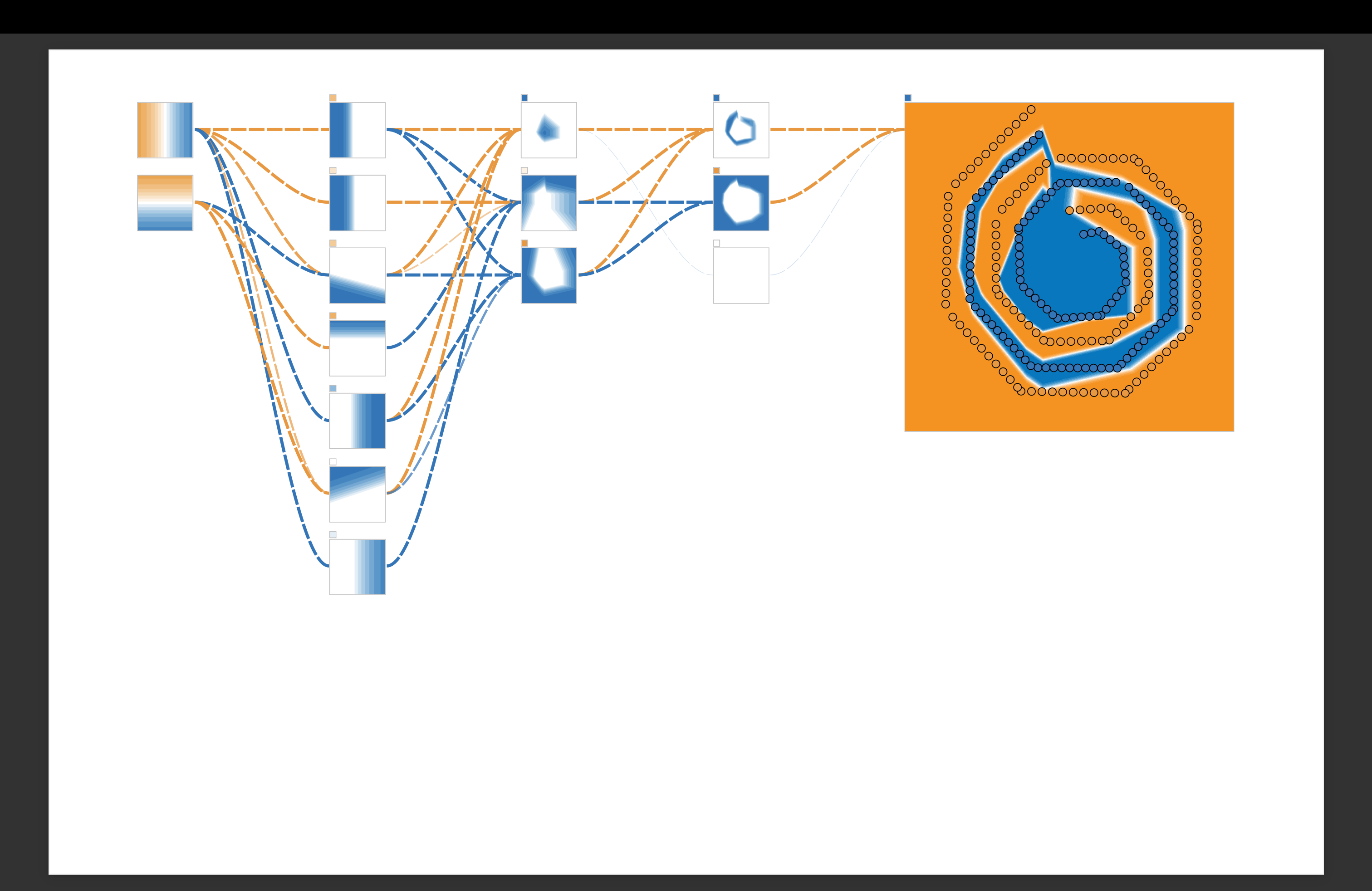}
        \caption{The minimal model found through the second phase of the combinatorial search that achieved over $99.5\%$ accuracy. The model has 45 nonzero parameters and an accuracy of $99.59\%$.}
        \label{fig:bench_733}
\end{figure}

\subsection{Analysis of Sparse Models}
Several observation can be deduced from the minimal models presented in Figures \ref{fig:bench_333} and \ref{fig:bench_733}:
\begin{description}
\item[Selective Connectivity] Both models exhibit selective connectivity and do not form connections with every neuron in the preceding layer. This suggests a more refined and efficient architectural design of the network.
\item[Edges to Spiral] The initial layers predominantly capture the spiral's edges. As we move deeper into the network, these edges are progressively integrated, forming polygonal shapes. In the last layers, these polygons are subtracted from one another to, roughly, reconstruct the spiral structure.
\item[Suboptimal Sparsity] The bottom neurons in the second and third layer of the model in Figure \ref{fig:bench_333}, and the bottom neuron in the third layer of the model in Figure \ref{fig:bench_733}, can be pruned to obtain a sparser model without significantly impacting the prediction of either model. Hence, although the models are sparse, they can be further pruned. We comment on this further in Sections \ref{sec:free_sparsity} and \ref{sec:prune_after}. 
\end{description}

\section{Pruning Algorithms Versus Combinatorial Search}
Given the results of the combinatorial search, we run the pruning experiments with the following budgets of nonzero weights:
\begin{gather*}
\{15, 16, 17, 18, 19,20, 21, 22, 23, 24,
25, 26, \\ 30, 33, 37, 40, 44, 50, 53, 55, 57, 60, 65\}.\end{gather*}
Figure \ref{fig:acc_vs_nnz_16} below shows the results. 

\begin{figure}[h]
        \centering
        \includegraphics[width=0.9\linewidth]{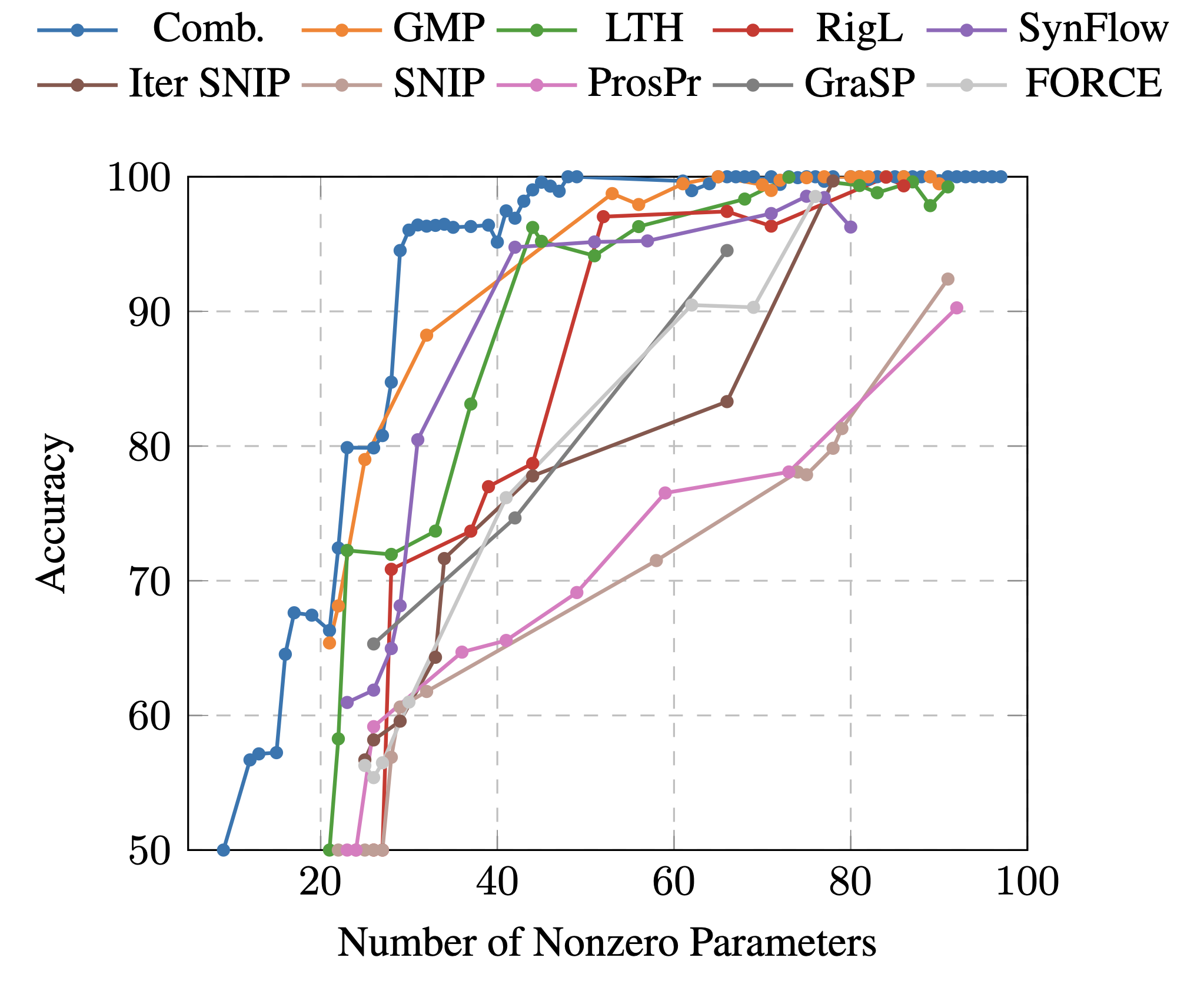}
        \caption{\textbf{Suboptimality of Pruning Algorithms.} The accuracy versus the number of nonzero parameters after training a four-layer, 16-width MLP on the Cubist Spiral dataset. A large gap is present across the board between the combinatorial search (Comb.) and the pruning algorithms, both in the $95\%-99.5\%$ accuracy range and for accuracies above $99.5\%$. The Pareto frontiers are manually extracted from a scatter plot, shown in Figure \ref{fig:scatter_acc_vs_nnz_16} in Appendix \ref{app:scatter}, that contains the accuracies of every pruning experiment.}
        \label{fig:acc_vs_nnz_16}
\end{figure}

\subsection{Suboptimality of Pruning Algorithms}
All pruning techniques suffer greatly in the $95\% - 99.5\%$ regime relative to the combinatorial search. In particular, the combinatorial search achieves above $95\%$ accuracy with just $30$ nonzeros. The second best is the dense-to-sparse method LTH requiring $44$ nonzeros to reach the accuracy threshold. The sparse-to-sparse method RigL reaches the threshold at $52$ nonzeros, while the sparsest pruning at initialization method that reaches the accuracy threshold is SynFlow with $51$ nonzeros. 

For accuracies above $99.5\%$, we again see a gap between pruning and the combinatorial search. The combinatorial search obtains above $99.5\%$ accuracy with $45$ nonzeros. The second-best method is GMP, which achieves a similar level of accuracy with $61$ nonzeros, while RigL reaches this accuracy threshold with $84$ nonzeros. The only pruning-at-initialization method that could reach the threshold within the tested nonzero budgets is Iter SNIP with $78$ nonzeros and $99.69\%$ accuracy.

\subsection{Visualization of Pruned Models}
\begin{figure}[h]
        \centering
        \includegraphics[width=\linewidth, trim=80 120 100 60, clip]{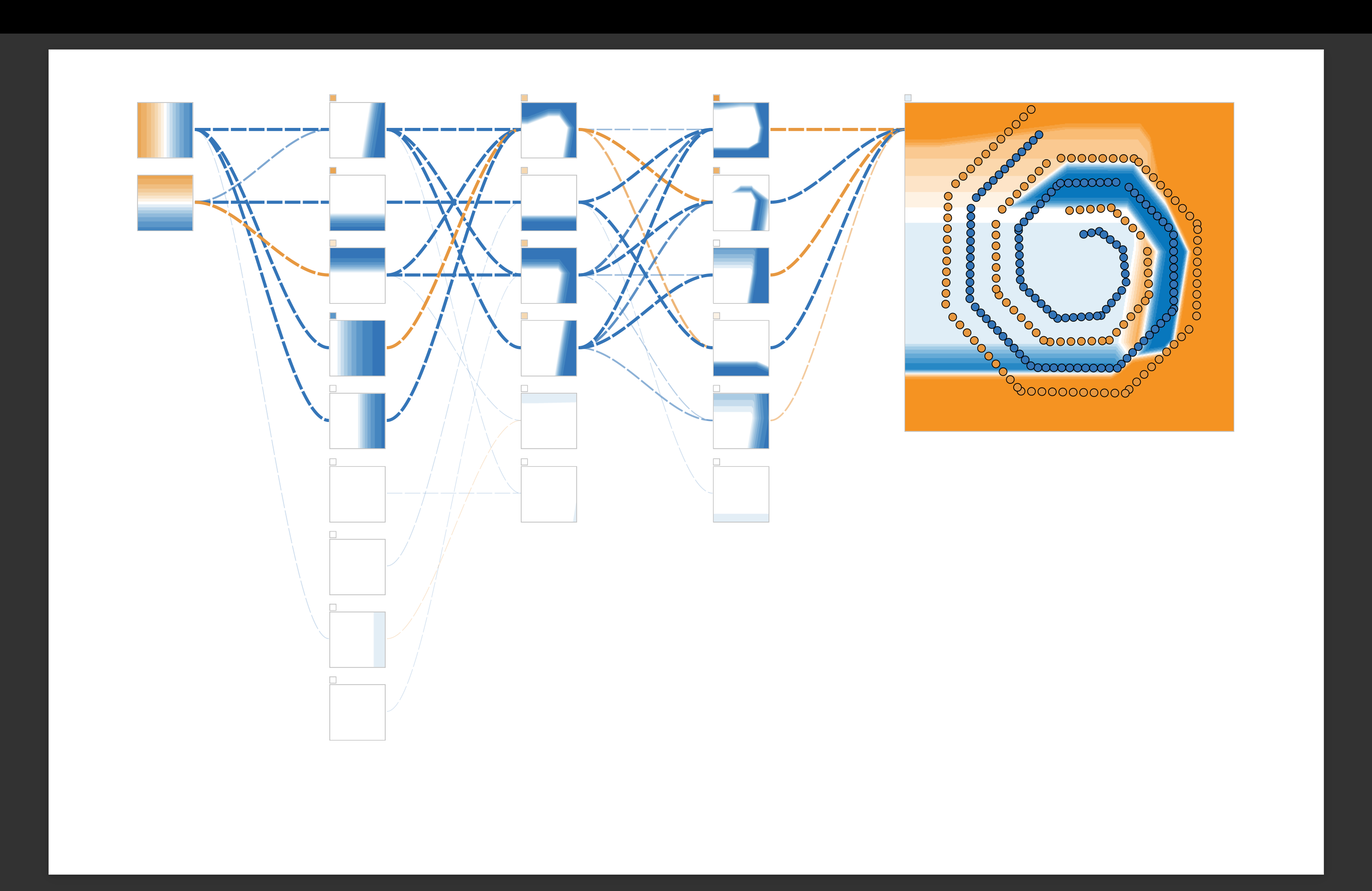}
        \caption{\textbf{ProsPr Suboptimality.} An example of a model, found through ProsPr, with $59$ nonzero parameters that attains an accuracy of $76.52\%$. The nonzero parameters attached to the bottom four neurons in layer one, bottom two neurons in layer two, and bottom neuron in layer three form multiple disconnected paths in the network.}
        \label{fig:prospr_example}
\end{figure}

To gain further insight as to why pruned models are struggling to match the combinatorial search, we visualize failed models generated by various pruning methods in Figures \ref{fig:prospr_example} and \ref{fig:gmp_example} below and comment on some key observations. Further examples are provided in Appendix \ref{app:more_viz}. 

\begin{figure}[h]
        \centering
        \includegraphics[width=\linewidth, trim=80 150 100 60, clip]{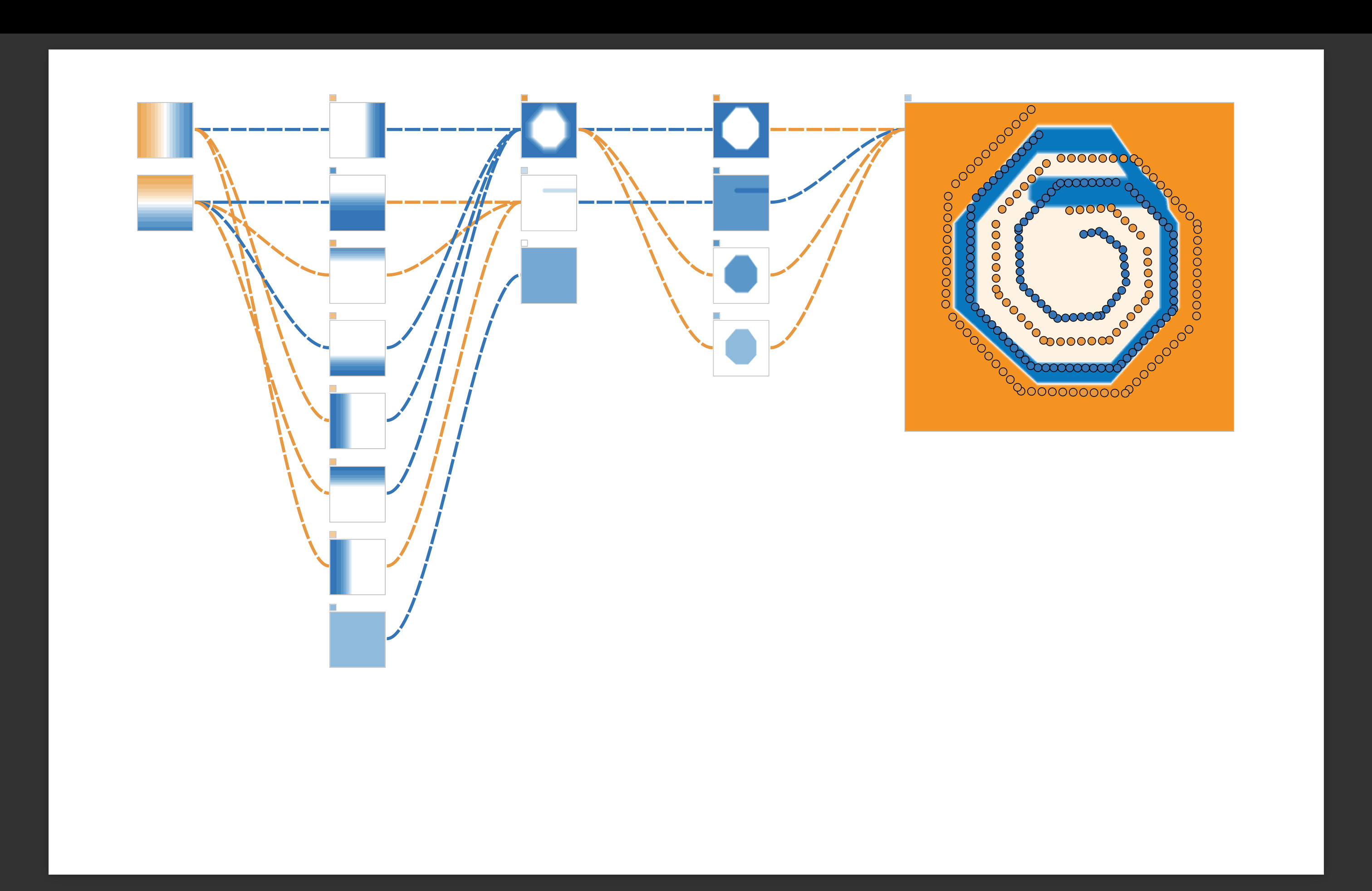}
        \caption{\textbf{GMP Suboptimality.} An example of a model, found through GMP, with $38$ nonzero parameters that attains an accuracy of $84.03\%$. The bottom neurons in layers two and three form a disconnected path. Magnitude-wise, no weights appear to be prunable.}
        \label{fig:gmp_example}
\end{figure}

\paragraph{Disconnected Paths} Current pruning algorithms are unable to properly align the weights between consecutive layers leading to \textit{disconnected paths}. This occurs when there is a path in the network that is either disconnected from the model input or output. Nonzero parameters in a disconnected path do not contribute to the expressiveness of the network and inflate the number of nonzeros in the model. 
\paragraph{Suboptimal Sparsity}\label{sec:free_sparsity} Similar to the models found by the combinatorial search, the model depicted in Figure \ref{fig:prospr_example} is foregoing a lot of sparsity that could be attained by magnitude pruning the model after training. This in part is due to the sub-optimal nature of current pruning algorithms requiring the final nonzero budget to be determined prior to any pruning or training being done.
\section{Impact of Overparameterization on Pruning} In theory, overparameterization should be beneficial to pruning as it increases the number of combinatorial options for sparsity masks from which to find the optimal mask for a sparse model. Contrary to this belief, Figure \ref{fig:diff_width} below shows the results from the experiments measuring the impact of overparameterization on the success of pruning algorithms. 
\begin{figure}[h]
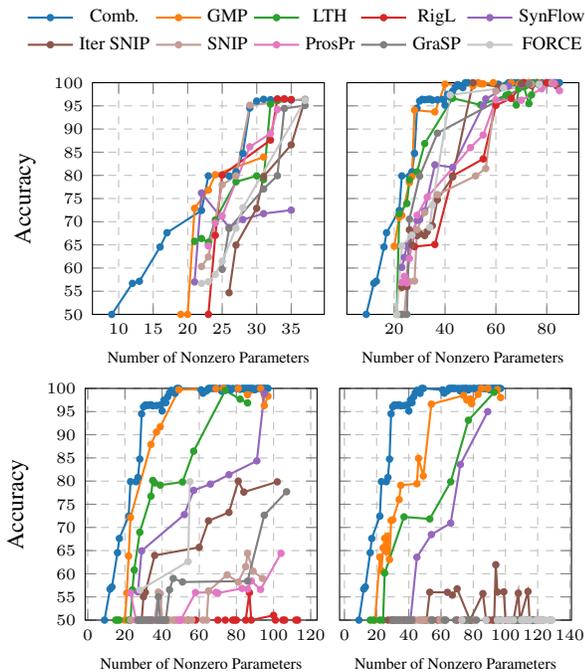

    \vspace{0.6cm}
    \centering
    \PlotTwoFigures{diff_width_3}{diff_width_7}{LastModelNNZ}{LastTestAcc}{true}
    \PlotTwoFigures{diff_width_64}{diff_width_256}{LastModelNNZ}{LastTestAcc}{false}
    \caption{\textbf{Overparameterization Impedes Pruning Algorithms.} The accuracy versus the number of nonzero parameters after training four-layer MLPs of varying widths on the Cubist Spiral dataset. From left to right then top to bottom, the subplots correspond to MLPs of width $3$, $7$, $64$, and $256$. More width variations are detailed in Appendix \ref{app:more_width}. The models obtained by the combinatorial search are included for reference.}
    \label{fig:diff_width}
\end{figure}
\subsection{Overparameterization Hinders Pruning} Figure \ref{fig:diff_width} shows that overparameterization harms the performance of most pruning techniques and that at width $256$, all pruning algorithms are largely failing. Furthermore, reducing the network width to $3$ increases the performance of current pruning algorithms indicating that current unstructured pruning approaches are inadequately performing structured pruning. In Appendix \ref{app:overparam_proof}, we further prove that overparameterization leads to more disconnected paths for pruning methods that utilize a random mask at initialization, like RigL.
\subsection{Optimal Width Limitations} Even when given the optimal width identified by the combinatorial search, the pruning methods are still unable to consistently match the accuracies that were shown to be empirically possible through the combinatorial search. Out of a total of \num{18954} experiments, only two instances of pruning were able to match or beat the combinatorial search: SNIP with 29 nonzeros and an accuracy of \(95.14\%\) and GMP with 40 nonzeros and an accuracy of \(99.68\%\). Both models were provided with the optimal widths of $3$ and $7$ respectively. 

\section{Impact of Initialization} \label{sec:fixed_init}
 In this section, we follow the experimental protocol detailed in Section \ref{sec:init_exper}, using the unmasked initialization of \textsc{Bench-995}. Analogous experiments for \textsc{Bench-95} are included in Appendix \ref{app:95_exper}.

\subsection{Combinatorial Search Using Optimal Initialization} 
 Running another round of the combinatorial search\footnote{Similar to the the (7,3,3) case, we do not perform an exhaustive combinatorial search over all the generated sparsity masks but rather only a subset. Details in Appendix \ref{app:995_subset}.} but this time training all models from the initialization of \textsc{Bench-995} recovers a sparser model that is visualized in Figure \ref{fig:bench_632} below. Due to the neuron configuration found (6,3,2), increasing the MLP width beyond 16 would not lead to the combinatorial search finding sparser solutions. Further explanation can be found in Appendix \ref{app:width_not_needed}.

 \begin{figure}[h]
        \centering
        \includegraphics[width=\linewidth, trim=80 340 100 60, clip]{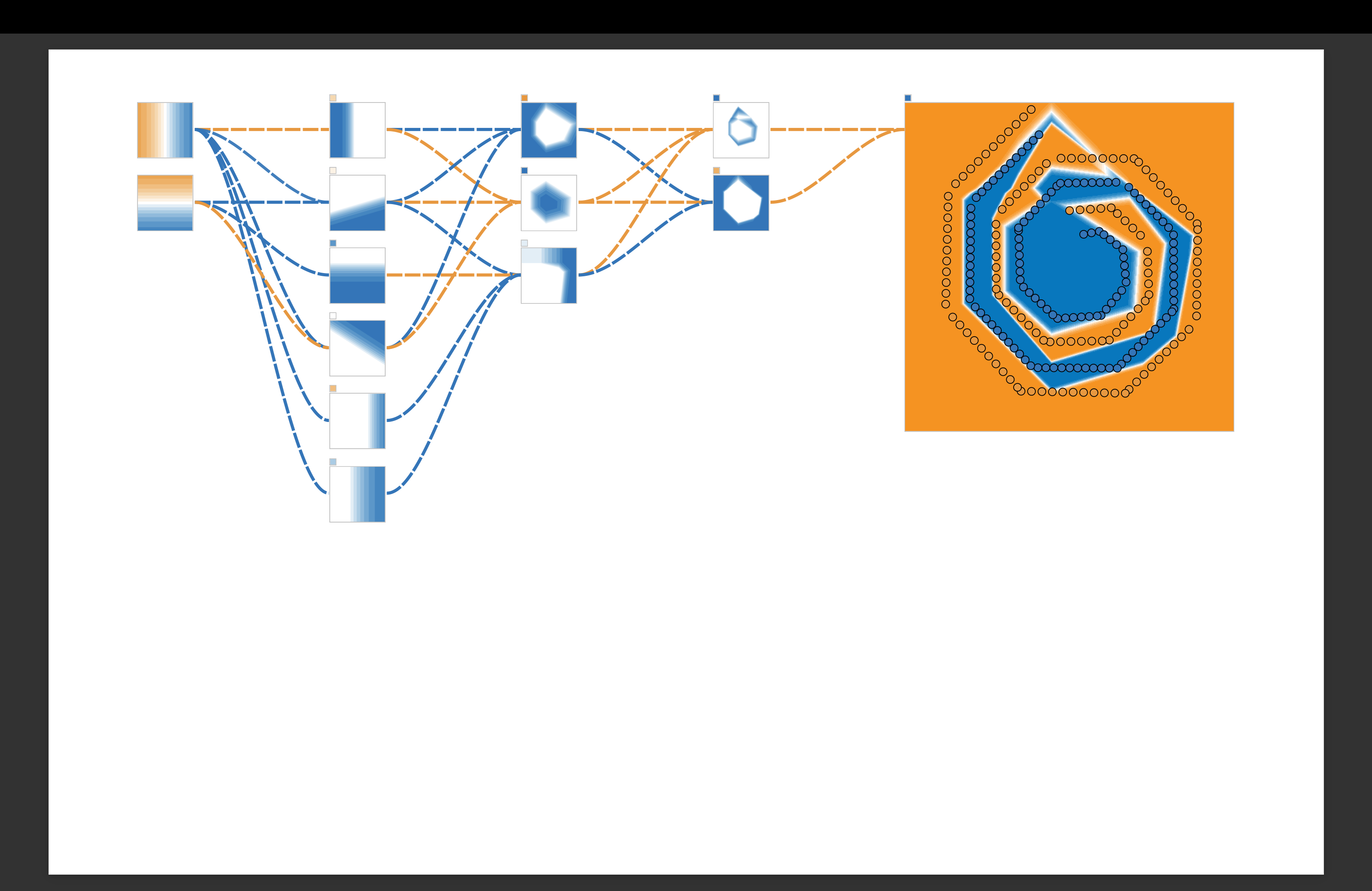}
        \caption{The minimal model found through the second run of the combinatorial search that achieved over $99.5\%$ accuracy. The number of neurons in each layer is (6,3,2). The model has 38 nonzero parameters and an accuracy of $99.70\%$.}
        \label{fig:bench_632}
\end{figure}

\subsection{Pruning after Training is Insufficient} \label{sec:prune_after}
Figure \ref{fig:bench_733} shows that pruning \textsc{Bench-995} obtained from the first combinatorial search after training will lead to a sparser model with a neuron configuration of $(7,3,2)$ and $42$ nonzeros. However, the second combinatorial search reveals a model that has a neuron configuration of $(6,3,2)$ containing \emph{less} nonzeros, $38$, than what would be obtained by pruning \textsc{Bench-995}. This indicates that magnitude pruning after training is insufficient to find a minimal model. 

\subsection{Pruning Fails with Optimal Initialization}\label{sec:SupportRecovery}
\begin{figure}[h]
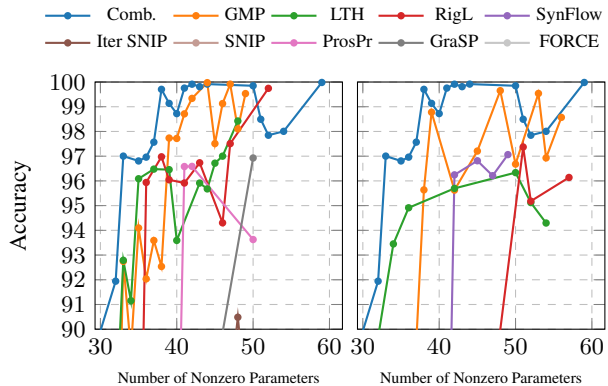

    \centering
    \vspace{0.6cm}    
    \PlotTwoFiguresHighAcc{995_recovery/recovery_6}{995_recovery/recovery_16}{Model NNZ1}{Test Acc1}{true}
    \caption{\textbf{Pruning Fails Under Optimal Conditions.} Models obtained by the combinatorial search and models that were pruned starting from the initialization of \textsc{Bench-995}. The pruned models in the left subplot were given the optimal structured sparsity mask of width 6 determined by the combinatorial search. The subplot on the right depicts models that were pruned directly from width 16.}
    \label{fig:recovery}
\end{figure}

Using the \textsc{Bench-995} initialization, Figure \ref{fig:recovery} below captures pruning's inability to recover a minimal sparsity mask for the model despite being given an ideal initialization. We can see that none of the pruning techniques are able to recover the minimal sparsity masks that the combinatorial search is able to find -- even when provided with the optimal width by masking the rows and columns of each layer down to the largest width of the weight matrices of the sparsest model found by the combinatorial search.

\begin{figure}[h]
        \begin{subfigure}[b]{\linewidth}
        \centering
        \includegraphics[width=\linewidth, trim=80 450 100 60, clip]{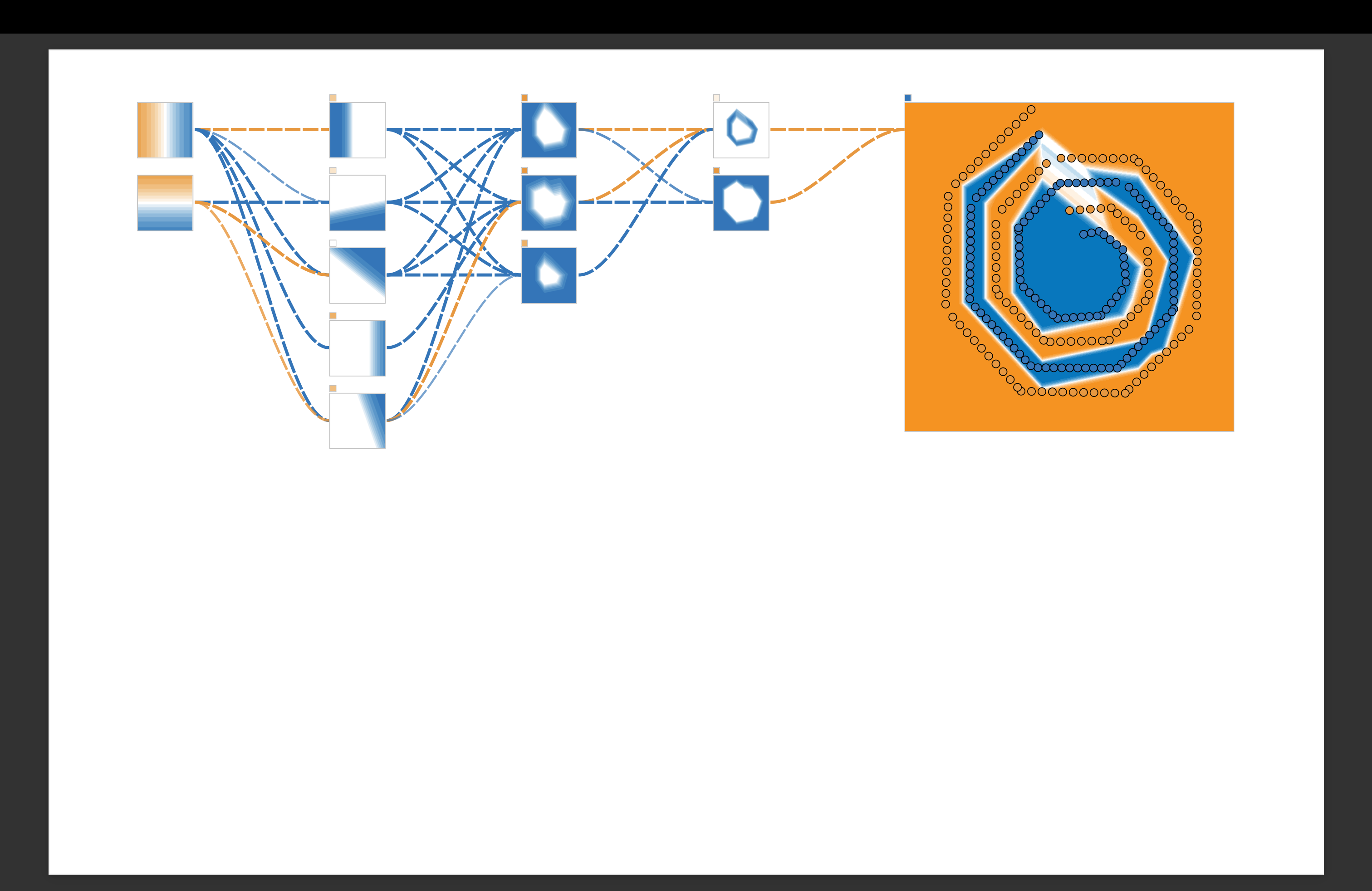}
        \caption{An example of a model, found through GMP, with $39$ nonzero parameters that attains an accuracy of $97.73\%$.}
        \label{fig:gmp_fix_init}
        \end{subfigure}
        \begin{subfigure}[b]{\linewidth}
        \centering
        \includegraphics[width=\linewidth, trim=80 400 100 60, clip]{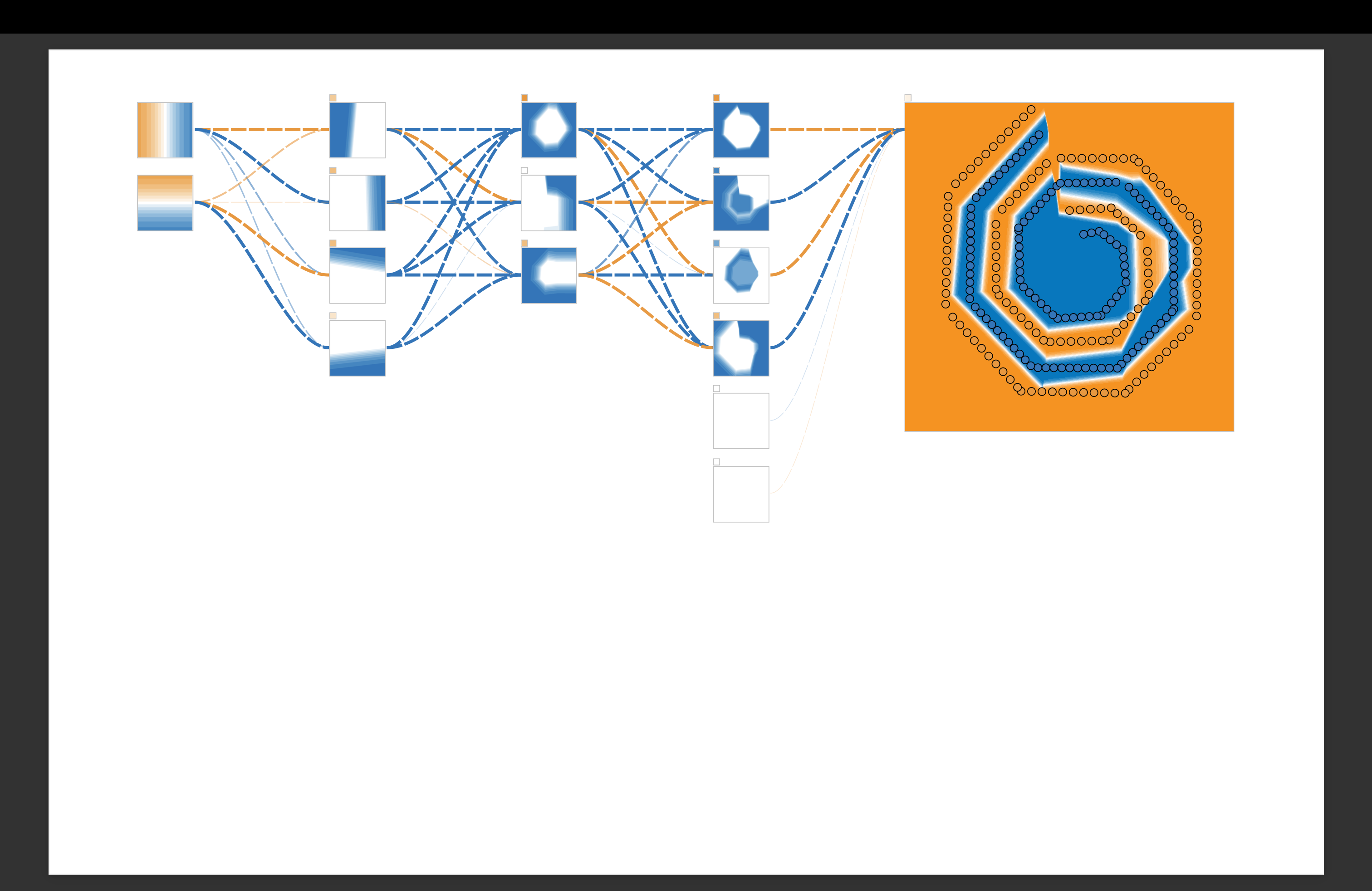}
        \caption{An example of a model, found through RigL, with $52$ nonzero parameters that attains an accuracy of $99.74\%$. There appear to be roughly eight nonzero parameters that could be pruned after training which would still lead to a model that contains more nonzeros than the one recovered by the combinatorial search depicted in Figure \ref{fig:bench_632}.}
        \label{fig:force_fix_init}
        \end{subfigure}
        \caption{\textbf{Pruning Algorithms Fail Under Optimal Conditions.} Despite being provided with the \textsc{Bench-995} initialization and the optimal width of 6, the pruned models depicted above still fall short of matching the models recovered by the combinatorial search.}
        \label{fig:failed_recovery}
\end{figure}

Depicted in Figure \ref{fig:failed_recovery} are models that were pruned with GMP and RigL that fail to match the model identified by the combinatorial search.

\section{Limitations of the Combinatorial Search}
\paragraph{Synthetic Datasets and Small Models} The high cost associated with performing the combinatorial search restricts our experiments to only synthetic datasets and small models. Still, the shortcomings that are already being exhibited by pruning techniques in such a simplistic task should raise concerns and inquiries into pruning’s current efficacy. While it is possible that the deficiencies of pruning algorithms observed do not extend to more complicated datasets and larger models, generally speaking, we would not expect an algorithm that does not work in a simple setting to work in a more complicated one.
\paragraph{No Guarantee of Sparsest} While our combinatorial search can find a model that is sparser than any of the models found by pruning, there is no guarantee that the combinatorial search finds the optimally sparse model. Rather the combinatorial search only provides a lower bound on the sparsity that is attainable for the four-layer MLP yet is not achieved by existing pruning algorithms – the sparsest models elude pruning.

\section{Related Works}
\paragraph{Sparse Representations and Compressed Sensing}
This work is predicated on the assumption that pruning algorithms ought to be able to identify the sparsest model. It is natural to question why such an assumption is even feasible. The rationale stems from both empirical and theoretical works in the fields of sparse representations and compressed sensing where it is known that, within the framework of linear models, if the underlying sparse solution is sufficiently sparse, then pruning algorithms will recover it. For further details, refer to \citep{donoho2006compressed,elad2010sparse, candes2005decoding, tropp2006just, tropp2004greed} and the cited literature.
\paragraph{Strong Tickets} Sparked by interest in the lottery ticket hypothesis \citep{frankle2018lottery}, numerous works have shown that, with high probability, there exists a subnetwork that can attain competitive performance within a sufficiently overparameterized randomly initialized network, called a \emph{strong lottery ticket} \citep{pmlr-v119-malach20a, ram_hidden, nips_log}. Our experiments reveal that, while the probability of identifying strong lottery tickets increases with the network's width, the effectiveness of current pruning methods in fact diminishes.
\paragraph{Random Pruning}
In line with our work, previous works have also benchmarked existing pruning techniques with naive pruning methods like random pruning \citep{liu2022the, galePruning}. What sets our work apart is that random pruning, much like existing pruning techniques, remains susceptible to disconnected paths. This poses a challenge in achieving the recovery of a maximally sparse model, especially at high levels of overparameterization. Our approach to the combinatorial search guarantees that misalignment between weights is impossible making the search significantly more efficient in finding a minimal model. 
\paragraph{Elucidating Pruning} Several recent studies have expressed concerns about the current state of pruning, particularly with inconsistent benchmarking. Both \citet{liu2023sparsity} and \citet{blalockState} proposed benchmarks for pruning, the former proposing SMC-Bench and the latter proposing ShrinkBench. \citet{frankle2021pruning} assessed several pruning at initialization techniques and remarked how they all perform similarly and are struggling to prune effectively at initialization. \citet{EvciGradFlow} showed that networks that were pruned at initialization have poor gradient flow leading to significantly worse generalization. For structured pruning, \citet{liu2018rethinking} observed that the common pipeline of fine-tuning the pruned model is, at best, comparable to just training the model from scratch and encouraged a more careful evaluation of structured pruning. We differentiate ourselves from prior works by comparing pruning against the sparsest of models, enabling us to underscore fundamental issues inherent in current pruning methods.
\paragraph{Plant 'n' Seek} In \citet{fischer2022plant}, the authors handcraft sparse networks to solve synthetic problems and plant them within a larger randomly initialized network. They find that current pruning techniques are unable to extract the sparse subnetwork from the larger network either at initialization or after training. Our work, on the other hand, does not require handcrafting sparse subnetworks and all training starts from a completely random initialization. This experimental setup is more representative of the standard pruning paradigm, where model sizes might not be large enough for strong tickets to exist at initialization with high probability.
\paragraph{Disconnected Paths} The tendency of pruning techniques to induce disconnected paths has previously been observed in prior works \citep{frankle2021pruning, JMLR:v24:22-0415, pham2023towards}. Both \citet{frankle2021pruning} and \citet{JMLR:v24:22-0415} propose measuring \emph{effective sparsity}, which accounts for the disconnected paths when assessing sparsity. In \citet{pham2023towards}, the authors found that the ratio of the number of connected paths to the number of active neurons in the model is crucial for the success of pruning (Node-Path Balancing Principle) and introduced a novel pruning method that maximizes both quantities.
\paragraph{Pruning and Layer-Collapse} It has been shown that current pruning techniques can inadvertently prune an entire layer at higher sparsity rates, effectively turning every path in the network into a disconnected one \citep{hayou2021robust}. In \citet{Lee2020A}, the authors showed that an initialization that preserves layerwise dynamical isometry can assist in preventing this while \citet{tanaka2020pruning} proposed the pruning technique SynFlow as a solution. Figure \ref{fig:diff_width} confirms that SynFlow is more robust to overparameterization compared to other pruning techniques but still fallible. 
Our work highlights that the problem is currently manifesting itself even at lower sparsity rates through disconnected paths and is more prevalent than just the catastrophic case where an entire layer is pruned.

\section{Conclusion}
We provided a comprehensive assessment of state-of-the-art pruning algorithms against the backdrop of ideal sparse networks obtained from a novel combinatorial search. Our findings reveal that current pruning algorithms fail to attain achievable sparsity levels -- even when given the optimal width and initialization. We associate this discrepancy with unstructured pruning's inadequacy at performing structured pruning, their failure to benefit from overparameterization, and their tendency to induce disconnected paths while also foregoing sparsity. Despite the simplicity of the dataset and network architectures employed in our study, we believe that the issues highlighted in our work are only exacerbated at larger scales and we hope that our methods and findings will be of assistance for future forays into the development of new pruning techniques.

\section*{Impact Statement}
This paper presents work whose goal is to advance the field of Machine Learning. There are many potential societal consequences of our work, none which we feel must be specifically highlighted here.

\section*{Acknowledgements}
We acknowledge the support of the Natural Sciences and Engineering Research Council of Canada (NSERC). This research was enabled in part by support provided by Compute Ontario (\url{https://www.computeontario.ca}) and the Digital Research Alliance of Canada (\url{https://alliancecan.ca/en}).

\bibliography{example_paper}

\begin{thebibliography}{42}
\providecommand{\natexlab}[1]{#1}
\providecommand{\url}[1]{\texttt{#1}}
\expandafter\ifx\csname urlstyle\endcsname\relax
  \providecommand{\doi}[1]{doi: #1}\else
  \providecommand{\doi}{doi: \begingroup \urlstyle{rm}\Url}\fi

\bibitem[Alizadeh et~al.(2021)Alizadeh, Tailor, Zintgraf, van Amersfoort, Farquhar, Lane, and Gal]{alizadeh2021prospect}
Alizadeh, M., Tailor, S.~A., Zintgraf, L.~M., van Amersfoort, J., Farquhar, S., Lane, N.~D., and Gal, Y.
\newblock Prospect pruning: Finding trainable weights at initialization using meta-gradients.
\newblock In \emph{International Conference on Learning Representations}, 2021.

\bibitem[Bellec et~al.(2018)Bellec, Kappel, Maass, and Legenstein]{bellec2018deep}
Bellec, G., Kappel, D., Maass, W., and Legenstein, R.
\newblock Deep rewiring: Training very sparse deep networks.
\newblock In \emph{International Conference on Learning Representations}, 2018.
\newblock URL \url{https://openreview.net/forum?id=BJ_wN01C-}.

\bibitem[Blalock et~al.(2020)Blalock, Ortiz, Frankle, and Guttag]{blalockState}
Blalock, D.~W., Ortiz, J. J.~G., Frankle, J., and Guttag, J.~V.
\newblock What is the state of neural network pruning?
\newblock In Dhillon, I.~S., Papailiopoulos, D.~S., and Sze, V. (eds.), \emph{MLSys}. mlsys.org, 2020.
\newblock URL \url{http://dblp.uni-trier.de/db/conf/mlsys/mlsys2020.html#BlalockOFG20}.

\bibitem[Candes \& Tao(2005)Candes and Tao]{candes2005decoding}
Candes, E.~J. and Tao, T.
\newblock Decoding by linear programming.
\newblock \emph{IEEE transactions on information theory}, 51\penalty0 (12):\penalty0 4203--4215, 2005.

\bibitem[de~Jorge et~al.(2021)de~Jorge, Sanyal, Behl, Torr, Rogez, and Dokania]{jorge2021progressive}
de~Jorge, P., Sanyal, A., Behl, H., Torr, P., Rogez, G., and Dokania, P.~K.
\newblock Progressive skeletonization: Trimming more fat from a network at initialization.
\newblock In \emph{International Conference on Learning Representations}, 2021.
\newblock URL \url{https://openreview.net/forum?id=9GsFOUyUPi}.

\bibitem[Dettmers \& Zettlemoyer(2019)Dettmers and Zettlemoyer]{dettmers2019snfs}
Dettmers, T. and Zettlemoyer, L.
\newblock Sparse networks from scratch: Faster training without losing performance.
\newblock \emph{CoRR}, abs/1907.04840, 2019.
\newblock URL \url{http://arxiv.org/abs/1907.04840}.

\bibitem[Donoho(2006)]{donoho2006compressed}
Donoho, D.~L.
\newblock Compressed sensing.
\newblock \emph{IEEE Transactions on information theory}, 52\penalty0 (4):\penalty0 1289--1306, 2006.

\bibitem[Elad(2010)]{elad2010sparse}
Elad, M.
\newblock \emph{Sparse and redundant representations: from theory to applications in signal and image processing}, volume~2.
\newblock Springer, 2010.

\bibitem[Evci et~al.(2020{\natexlab{a}})Evci, Gale, Menick, Castro, and Elsen]{evci2020rigging}
Evci, U., Gale, T., Menick, J., Castro, P.~S., and Elsen, E.
\newblock Rigging the lottery: Making all tickets winners.
\newblock In \emph{International Conference on Machine Learning}, pp.\  2943--2952. PMLR, 2020{\natexlab{a}}.

\bibitem[Evci et~al.(2020{\natexlab{b}})Evci, Ioannou, Keskin, and Dauphin]{EvciGradFlow}
Evci, U., Ioannou, Y.~A., Keskin, C., and Dauphin, Y.~N.
\newblock Gradient flow in sparse neural networks and how lottery tickets win.
\newblock \emph{CoRR}, abs/2010.03533, 2020{\natexlab{b}}.
\newblock URL \url{https://arxiv.org/abs/2010.03533}.

\bibitem[Fischer \& Burkholz(2022)Fischer and Burkholz]{fischer2022plant}
Fischer, J. and Burkholz, R.
\newblock Plant 'n' seek: Can you find the winning ticket?
\newblock In \emph{International Conference on Learning Representations}, 2022.
\newblock URL \url{https://openreview.net/forum?id=9n9c8sf0xm}.

\bibitem[Frankle \& Carbin(2018)Frankle and Carbin]{frankle2018lottery}
Frankle, J. and Carbin, M.
\newblock The lottery ticket hypothesis: Finding sparse, trainable neural networks.
\newblock In \emph{International Conference on Learning Representations}, 2018.

\bibitem[Frankle et~al.(2021)Frankle, Dziugaite, Roy, and Carbin]{frankle2021pruning}
Frankle, J., Dziugaite, G.~K., Roy, D., and Carbin, M.
\newblock Pruning neural networks at initialization: Why are we missing the mark?
\newblock In \emph{International Conference on Learning Representations}, 2021.
\newblock URL \url{https://openreview.net/forum?id=Ig-VyQc-MLK}.

\bibitem[Gale et~al.(2019)Gale, Elsen, and Hooker]{galePruning}
Gale, T., Elsen, E., and Hooker, S.
\newblock The state of sparsity in deep neural networks.
\newblock \emph{CoRR}, abs/1902.09574, 2019.
\newblock URL \url{http://arxiv.org/abs/1902.09574}.

\bibitem[Han et~al.(2015)Han, Pool, Tran, and Dally]{han2015learning}
Han, S., Pool, J., Tran, J., and Dally, W.
\newblock Learning both weights and connections for efficient neural network.
\newblock \emph{Advances in neural information processing systems}, 28, 2015.

\bibitem[Hassibi \& Stork(1992)Hassibi and Stork]{hassibi1992second}
Hassibi, B. and Stork, D.
\newblock Second order derivatives for network pruning: Optimal brain surgeon.
\newblock \emph{Advances in neural information processing systems}, 5, 1992.

\bibitem[Hassibi et~al.(1993)Hassibi, Stork, and Wolff]{hassibi1993optimal}
Hassibi, B., Stork, D.~G., and Wolff, G.~J.
\newblock Optimal brain surgeon and general network pruning.
\newblock In \emph{IEEE international conference on neural networks}, pp.\  293--299. IEEE, 1993.

\bibitem[Hayou et~al.(2021)Hayou, Ton, Doucet, and Teh]{hayou2021robust}
Hayou, S., Ton, J.-F., Doucet, A., and Teh, Y.~W.
\newblock Robust pruning at initialization.
\newblock In \emph{International Conference on Learning Representations}, 2021.
\newblock URL \url{https://openreview.net/forum?id=vXj_ucZQ4hA}.

\bibitem[Hoefler et~al.(2021)Hoefler, Alistarh, Ben-Nun, Dryden, and Peste]{heofler_prune_survey}
Hoefler, T., Alistarh, D., Ben-Nun, T., Dryden, N., and Peste, A.
\newblock Sparsity in deep learning: Pruning and growth for efficient inference and training in neural networks.
\newblock \emph{Journal of Machine Learning Research}, 22\penalty0 (241):\penalty0 1--124, 2021.
\newblock URL \url{http://jmlr.org/papers/v22/21-0366.html}.

\bibitem[LeCun et~al.(1989)LeCun, Denker, and Solla]{lecun1989optimal}
LeCun, Y., Denker, J., and Solla, S.
\newblock Optimal brain damage.
\newblock \emph{Advances in neural information processing systems}, 2, 1989.

\bibitem[Lee et~al.(2019)Lee, Ajanthan, and Torr]{lee2019snip}
Lee, N., Ajanthan, T., and Torr, P.
\newblock Snip: single-shot network pruning based on connection sensitivity.
\newblock In \emph{International Conference on Learning Representations}. Open Review, 2019.

\bibitem[Lee et~al.(2020)Lee, Ajanthan, Gould, and Torr]{Lee2020A}
Lee, N., Ajanthan, T., Gould, S., and Torr, P. H.~S.
\newblock A signal propagation perspective for pruning neural networks at initialization.
\newblock In \emph{International Conference on Learning Representations}, 2020.
\newblock URL \url{https://openreview.net/forum?id=HJeTo2VFwH}.

\bibitem[Li et~al.(2017)Li, Kadav, Durdanovic, Samet, and Graf]{li2017pruning}
Li, H., Kadav, A., Durdanovic, I., Samet, H., and Graf, H.~P.
\newblock Pruning filters for efficient convnets.
\newblock In \emph{International Conference on Learning Representations}, 2017.
\newblock URL \url{https://openreview.net/forum?id=rJqFGTslg}.

\bibitem[Liu et~al.(2022)Liu, Chen, Chen, Shen, Mocanu, Wang, and Pechenizkiy]{liu2022the}
Liu, S., Chen, T., Chen, X., Shen, L., Mocanu, D.~C., Wang, Z., and Pechenizkiy, M.
\newblock The unreasonable effectiveness of random pruning: Return of the most naive baseline for sparse training.
\newblock In \emph{International Conference on Learning Representations}, 2022.
\newblock URL \url{https://openreview.net/forum?id=VBZJ_3tz-t}.

\bibitem[Liu et~al.(2023)Liu, Chen, Zhang, Chen, Huang, JAISWAL, and Wang]{liu2023sparsity}
Liu, S., Chen, T., Zhang, Z., Chen, X., Huang, T., JAISWAL, A.~K., and Wang, Z.
\newblock Sparsity may cry: Let us fail (current) sparse neural networks together!
\newblock In \emph{The Eleventh International Conference on Learning Representations}, 2023.
\newblock URL \url{https://openreview.net/forum?id=J6F3lLg4Kdp}.

\bibitem[Liu et~al.(2019)Liu, Sun, Zhou, Huang, and Darrell]{liu2018rethinking}
Liu, Z., Sun, M., Zhou, T., Huang, G., and Darrell, T.
\newblock Rethinking the value of network pruning.
\newblock In \emph{International Conference on Learning Representations}, 2019.
\newblock URL \url{https://openreview.net/forum?id=rJlnB3C5Ym}.

\bibitem[Luo et~al.(2017)Luo, Wu, and Lin]{Luo_2017_ICCV}
Luo, J.-H., Wu, J., and Lin, W.
\newblock Thinet: A filter level pruning method for deep neural network compression.
\newblock In \emph{Proceedings of the IEEE International Conference on Computer Vision (ICCV)}, Oct 2017.

\bibitem[Malach et~al.(2020)Malach, Yehudai, Shalev-Schwartz, and Shamir]{pmlr-v119-malach20a}
Malach, E., Yehudai, G., Shalev-Schwartz, S., and Shamir, O.
\newblock Proving the lottery ticket hypothesis: Pruning is all you need.
\newblock In III, H.~D. and Singh, A. (eds.), \emph{Proceedings of the 37th International Conference on Machine Learning}, volume 119 of \emph{Proceedings of Machine Learning Research}, pp.\  6682--6691. PMLR, 13--18 Jul 2020.
\newblock URL \url{https://proceedings.mlr.press/v119/malach20a.html}.

\bibitem[Mocanu et~al.(2018)Mocanu, Mocanu, Stone, Nguyen, Gibescu, and Liotta]{mocanu_scalable_2018}
Mocanu, D.~C., Mocanu, E., Stone, P., Nguyen, P.~H., Gibescu, M., and Liotta, A.
\newblock Scalable training of artificial neural networks with adaptive sparse connectivity inspired by network science.
\newblock \emph{Nature Communications}, 9\penalty0 (1):\penalty0 2383, June 2018.
\newblock ISSN 2041-1723.
\newblock \doi{10.1038/s41467-018-04316-3}.
\newblock URL \url{https://doi.org/10.1038/s41467-018-04316-3}.

\bibitem[Mostafa \& Wang(2019)Mostafa and Wang]{pmlr-v97-mostafa19a}
Mostafa, H. and Wang, X.
\newblock Parameter efficient training of deep convolutional neural networks by dynamic sparse reparameterization.
\newblock In Chaudhuri, K. and Salakhutdinov, R. (eds.), \emph{Proceedings of the 36th International Conference on Machine Learning}, volume~97 of \emph{Proceedings of Machine Learning Research}, pp.\  4646--4655. PMLR, 09--15 Jun 2019.
\newblock URL \url{https://proceedings.mlr.press/v97/mostafa19a.html}.

\bibitem[Orseau et~al.(2020)Orseau, Hutter, and Rivasplata]{nips_log}
Orseau, L., Hutter, M., and Rivasplata, O.
\newblock Logarithmic pruning is all you need.
\newblock In Larochelle, H., Ranzato, M., Hadsell, R., Balcan, M., and Lin, H. (eds.), \emph{Advances in Neural Information Processing Systems}, volume~33, pp.\  2925--2934. Curran Associates, Inc., 2020.
\newblock URL \url{https://proceedings.neurips.cc/paper_files/paper/2020/file/1e9491470749d5b0e361ce4f0b24d037-Paper.pdf}.

\bibitem[Pham et~al.(2023)Pham, Ta, Liu, Xiang, Le, Wen, and Tran-Thanh]{pham2023towards}
Pham, H., Ta, T.-A., Liu, S., Xiang, L., Le, D.~D., Wen, H., and Tran-Thanh, L.
\newblock Towards data-agnostic pruning at initialization: What makes a good sparse mask?
\newblock In \emph{Thirty-seventh Conference on Neural Information Processing Systems}, 2023.
\newblock URL \url{https://openreview.net/forum?id=xdOoCWCYaY}.

\bibitem[Ramanujan et~al.(2020)Ramanujan, Wortsman, Kembhavi, Farhadi, and Rastegari]{ram_hidden}
Ramanujan, V., Wortsman, M., Kembhavi, A., Farhadi, A., and Rastegari, M.
\newblock What’s hidden in a randomly weighted neural network?
\newblock In \emph{2020 IEEE/CVF Conference on Computer Vision and Pattern Recognition (CVPR)}, pp.\  11890--11899, 2020.
\newblock \doi{10.1109/CVPR42600.2020.01191}.

\bibitem[Smilkov et~al.(2017)Smilkov, Carter, Sculley, Vi{\'e}gas, and Wattenberg]{smilkov2017direct}
Smilkov, D., Carter, S., Sculley, D., Vi{\'e}gas, F.~B., and Wattenberg, M.
\newblock Direct-manipulation visualization of deep networks.
\newblock \emph{arXiv preprint arXiv:1708.03788}, 2017.

\bibitem[Tanaka et~al.(2020)Tanaka, Kunin, Yamins, and Ganguli]{tanaka2020pruning}
Tanaka, H., Kunin, D., Yamins, D.~L., and Ganguli, S.
\newblock Pruning neural networks without any data by iteratively conserving synaptic flow.
\newblock \emph{Advances in neural information processing systems}, 33:\penalty0 6377--6389, 2020.

\bibitem[Tropp(2004)]{tropp2004greed}
Tropp, J.~A.
\newblock Greed is good: Algorithmic results for sparse approximation.
\newblock \emph{IEEE Transactions on Information theory}, 50\penalty0 (10):\penalty0 2231--2242, 2004.

\bibitem[Tropp(2006)]{tropp2006just}
Tropp, J.~A.
\newblock Just relax: Convex programming methods for identifying sparse signals in noise.
\newblock \emph{IEEE transactions on information theory}, 52\penalty0 (3):\penalty0 1030--1051, 2006.

\bibitem[Vysogorets \& Kempe(2023)Vysogorets and Kempe]{JMLR:v24:22-0415}
Vysogorets, A. and Kempe, J.
\newblock Connectivity matters: Neural network pruning through the lens of effective sparsity.
\newblock \emph{Journal of Machine Learning Research}, 24\penalty0 (99):\penalty0 1--23, 2023.
\newblock URL \url{http://jmlr.org/papers/v24/22-0415.html}.

\bibitem[Wang et~al.(2019)Wang, Zhang, and Grosse]{wang2019picking}
Wang, C., Zhang, G., and Grosse, R.
\newblock Picking winning tickets before training by preserving gradient flow.
\newblock In \emph{International Conference on Learning Representations}, 2019.

\bibitem[Wen et~al.(2016)Wen, Wu, Wang, Chen, and Li]{Wen2016structure}
Wen, W., Wu, C., Wang, Y., Chen, Y., and Li, H.
\newblock Learning structured sparsity in deep neural networks.
\newblock In \emph{Proceedings of the 30th International Conference on Neural Information Processing Systems}, NIPS'16, pp.\  2082–2090, Red Hook, NY, USA, 2016. Curran Associates Inc.
\newblock ISBN 9781510838819.

\bibitem[Zhou et~al.(2019)Zhou, Lan, Liu, and Yosinski]{zhou_2019_dlt}
Zhou, H., Lan, J., Liu, R., and Yosinski, J.
\newblock Deconstructing lottery tickets: Zeros, signs, and the supermask.
\newblock In \emph{Advances in Neural Information Processing Systems}, 2019.

\bibitem[Zhu \& Gupta(2018)Zhu and Gupta]{zhuGMP}
Zhu, M.~H. and Gupta, S.
\newblock To prune, or not to prune: Exploring the efficacy of pruning for model compression, 2018.
\newblock URL \url{https://openreview.net/forum?id=S1lN69AT-}.

\end{thebibliography}
\bibliographystyle{icml2024}

 \newpage
 \appendix
 \onecolumn
 \section{\textsc{Bench-95} Initialization Experiments}\label{app:95_exper}
We run the same experiments as detailed in Section \ref{sec:fixed_init} but using the \textsc{Bench-95} initialization. 

\begin{figure}[h]
        \centering
        \includegraphics[width=\linewidth, trim=80 500 100 60, clip]{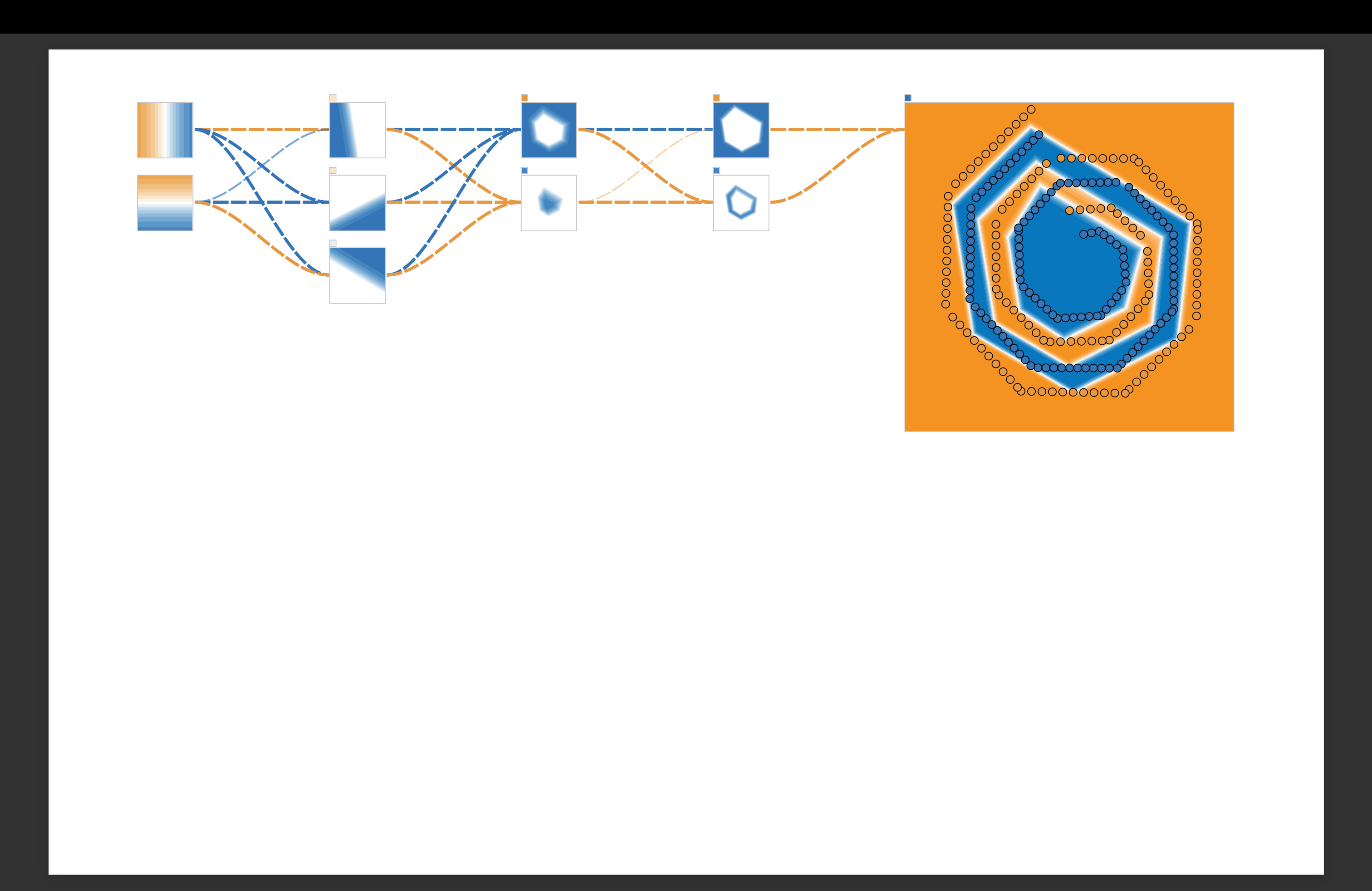}
        \caption{Benchmark Model with 26 nonzeros that was trained using the \textsc{Bench-95} initialization. Attains $96.47\%$ accuracy.}
        \label{fig:bench_322}
\end{figure}

 Starting from the fixed initialization of \textsc{Bench-95}, the combinatorial search is now able to identify a minimal neuron configuration of (3,2,2) and a benchmark model with 26 nonzeros that attains $96.47\%$ accuracy.

 Figure \ref{fig:recovery_95} below depicts models obtained by the combinatorial search and models that were pruned starting from the \textsc{Bench-95} initialization.

\begin{figure}[!h]
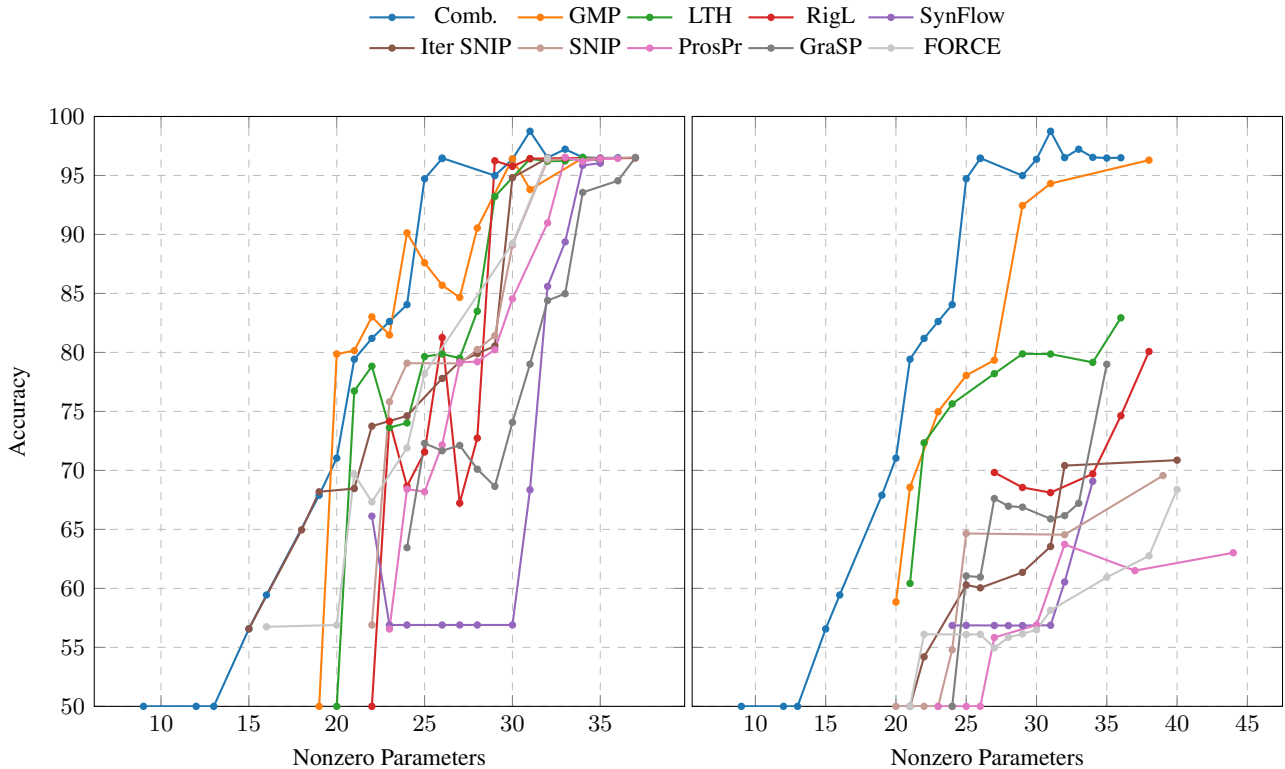

    \centering
    \PlotTwoFiguresApp{95_recovery/recovery_3}{95_recovery/recovery_16}{LastNNZ}{LastTest}{true}
    \caption{Analogous plot to Figure \ref{fig:recovery} but models are trained from the BENCH-95 initialization. The left subplot shows models that were provided with the optimal width of 3 via a structured sparsity mask. The right subplot shows models that were directly pruned from a width of 16. The scatter plot used to generate the Pareto frontiers in this plot are depicted in Figure \ref{fig:scatter_recover_95}.}
    \label{fig:recovery_95}
\end{figure}

\section{Hyperparameters for Pruning Algorithms}\label{app:prune_hyper}
    To enforce sparsity in the models, we base our implementation on code from: \url{https://github.com/facebookresearch/open_lth}.
 \subsection{Hyperparameters for GMP}
 We use the cubic decay schedule detailed in \citep{zhuGMP} with 199 pruning steps spread out evenly across the 50 epochs of training. We base our implementation of magnitude pruning on code from: \url{https://github.com/facebookresearch/open_lth}.
  \subsection{Hyperparameters for LTH}
  We divide the training into five different blocks of 50 epochs of training (i.e. 250 epochs total). Between each block, we prune $p\%$ of the weights where $p$ is chosen so that the final sparsity is reached. Then, we rewind the model back to initialization along with the learning rate scheduler. This ensures that the hyperparameters are consistent with the rest of the experiments. 
 \subsection{Hyperparameters for RigL}
 We use the ERK distribution to determine the sparsity for each layer. For the update schedule, we utilize the hyperparameters: $\Delta T = 200, \alpha = 0.3$, $f_{decay}$ to be cosine annealing, and we stop updating the mask $75\%$ through training. We base our implementation of RigL on code from: \url{https://github.com/verbose-avocado/rigl-torch}
 and the implementation of the ERK distribution on code from: \url{https://github.com/google-research/rigl}.
  \subsection{Hyperparameters for ProsPr}
 We use the momentum and learning rate used in training to calculate the meta-gradients. We also perform three training steps to calculate the meta-gradients. We base our implementation of ProsPr on code from: \url{https://github.com/mil-ad/prospr}.
 \subsection{Hyperparameters for GraSP}
 We use the hyperparameters and base our implementation of GraSP on code from: \url{https://github.com/alecwangcq/GraSP}.
\subsection{Hyperparameters for Iter SNIP and FORCE}
  We set the number of iterations to be 10 using the exponential decay schedule and just one batch to compute the average saliency per iteration. We base our implementations of Iter SNIP and FORCE on code from: \url{https://github.com/naver/force}.
\subsection{Hyperparameters for SynFlow}
We set the number of iterations to 100 with an exponential pruning schedule.  We base our implementation of SynFlow on code from: \url{https://github.com/ganguli-lab/Synaptic-Flow}.
\subsection{Implementation of SNIP}
 We base our implementation of SNIP on code from: \url{https://github.com/mil-ad/snip}.

 \section{FLOPS Measurements}\label{app:flops}
 \PlotBar

 The FLOPS required for each pruning technique as depicted in Table \ref{table:prune_type} are measured by pruning and training an MLP of width 16 with 55 nonzero weights. We measure the FLOPS by computing the number of floating operations utilized by the nonzero parameters (including biases) in the model throughout pruning and training. We base our implementation on code from: \url{https://github.com/simochen/model-tools}
 \section{Selection of Subsets of Unstructured Masks for Combinatorial Search}\label{app:995_subset}
 We restrict the unstructured masks from two fronts. The first front is by restricting the masks of $\mW^{[2]}$ to a subset of all possible masks. This is done through the for loop on line 3 in Algorithm \ref{alg:combinatorial} and instead of choosing all eligible unstructured masks, we simply choose the first three for the (7,3,3) configuration that are deemed eligible. For the (6,3,2) configuration, we choose the first six eligible unstructured masks. The second front is by restricting to the set of unstructured masks for the model to those that contain fewer nonzeros than a certain amount. This amount was set to 49 for the (7,3,3) configuration and 45 for the (6,3,2) configuration. We also run these experiments with a single learning rate of $0.05$. 
 
 \section{Dependence on Initialization}\label{app:dep_init}
 The combinatorial search does not achieve the same sparse network if different initializations are provided. A similar dependence of the mask on initialization has previously been observed in \citet{frankle2018lottery} and studied in \citet{zhou_2019_dlt}. In the latter, the authors demonstrated that the mask only depends on the sign pattern of the initialization. Inspired by such observations, we ran exploratory experiments that indicated otherwise, at least in the case of the combinatorial search. As highlighted in \citet{EvciGradFlow}, the lack of consistency across initializations might be attributed to poor gradient flow and overall challenges associated with training a sparse network from scratch.

\section{Scatter Plots}\label{app:scatter}
\begin{figure*}

    \centering
    \includegraphics[width=0.9\linewidth]{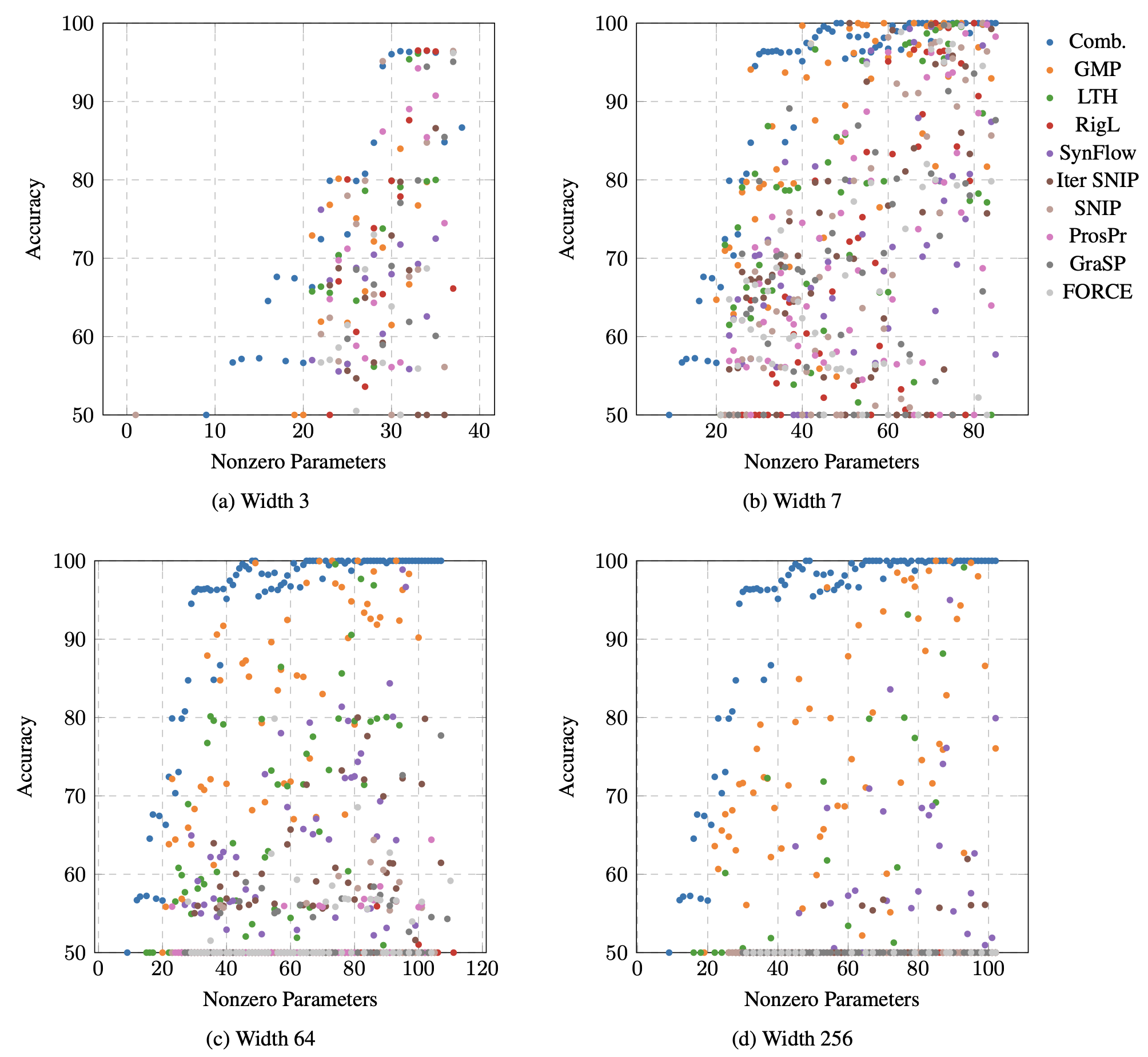}
    \caption{Scatter plots containing the best performing run for each pruning approach at a given number of nonzeros in the model. Pareto frontiers depicted in Figure \ref{fig:diff_width} are interpolated from the datapoints depicted in this plot.}
    \label{fig:scatter_diff_width}
\end{figure*}

The scatter plots that were used to generate the Pareto frontiers in Figures \ref{fig:acc_vs_nnz_16}, \ref{fig:diff_width}, and \ref{fig:recovery} are depicted in Figures \ref{fig:scatter_acc_vs_nnz_16}, \ref{fig:scatter_diff_width}, and \ref{fig:scatter_recover_995} respectively. 

\section{Overparameterization: More Widths}\label{app:more_width}
\begin{figure}[h]
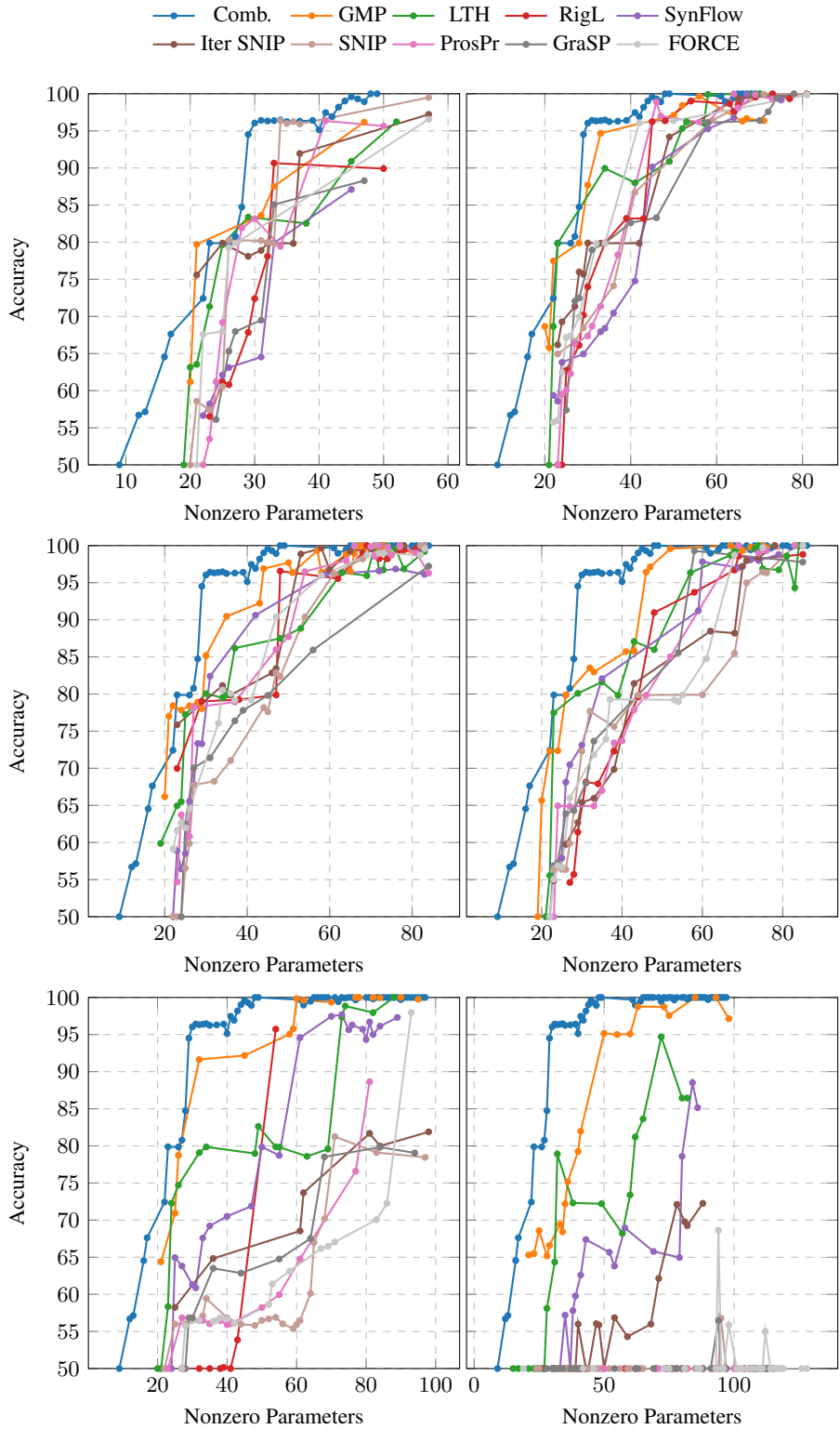

     \centering
     \PlotSixFigures{diff_width_4}{diff_width_5}{LastModelNNZ}{LastTestAcc}{true}
    
     \PlotSixFigures{diff_width_6}{diff_width_8}{LastModelNNZ}{LastTestAcc}{false}
    
     \PlotSixFigures{diff_width_32}{diff_width_128}{LastModelNNZ}{LastTestAcc}{false}
    \caption{\textbf{More Widths} The accuracy versus the number of nonzero parameters after training four-layer MLPs of varying widths on the Cubist Spiral dataset. From left to right then top to bottom, the subplots correspond to MLPs of width 4, 5, 6, 8, 32, and 128.}
    \label{fig:more_diff_width}
\end{figure}

\begin{figure*}
    \centering
    \includegraphics[width=0.9\linewidth]{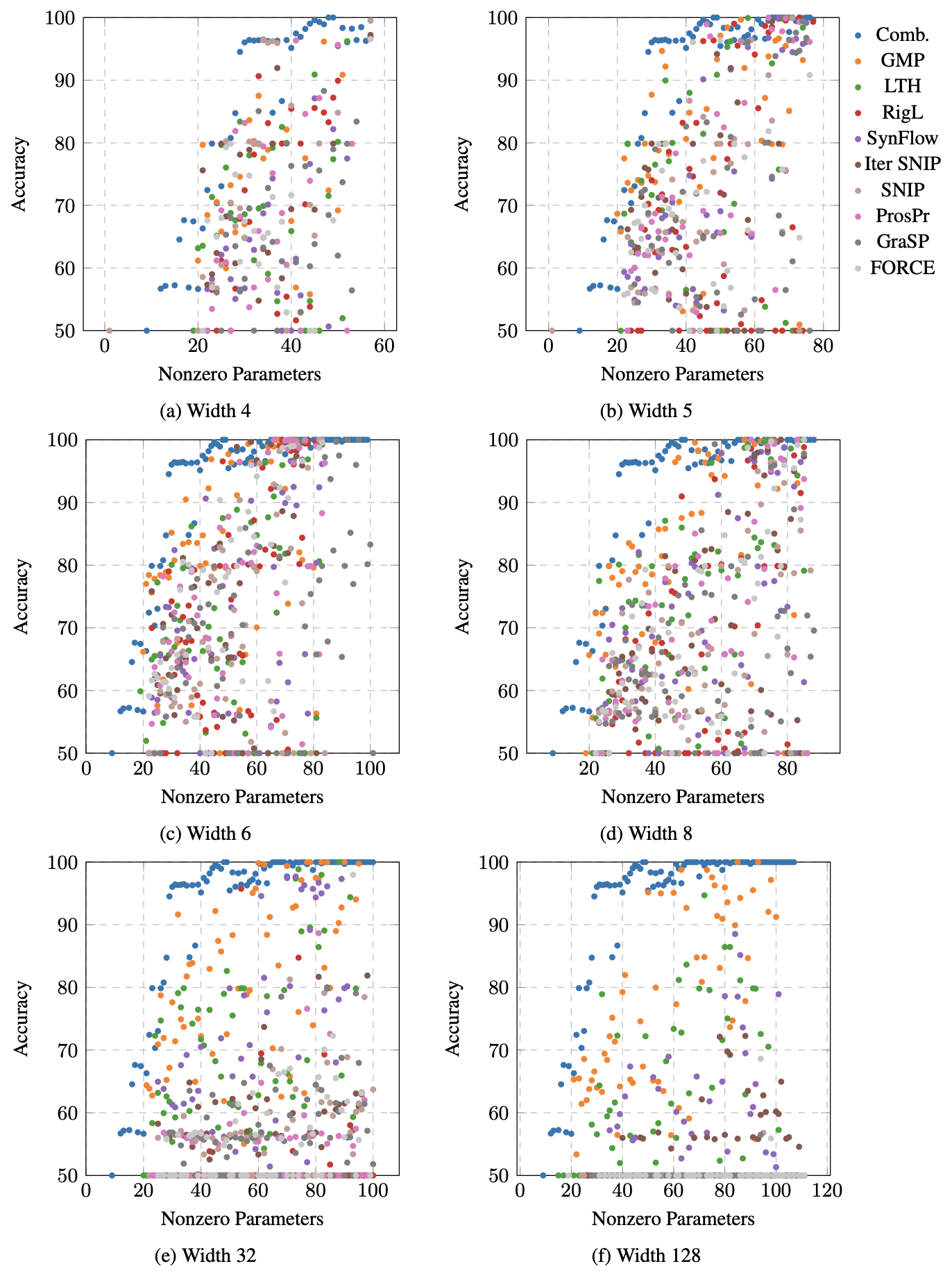}
    \caption{Scatter plots containing the best performing run for each pruning approach at a given number of nonzeros in the model. Pareto frontiers depicted in Figure \ref{fig:more_diff_width} are interpolated from the datapoints depicted in this plot.}
    \label{fig:scatter_more_diff_width}
\end{figure*}

We include the plots corresponding to widths 4,5,6,8, 32, and 128 in Figure \ref{fig:more_diff_width}. The scatter plots that were used to generate the Pareto frontiers in these plots are shown in Figure \ref{fig:scatter_more_diff_width}. 

\section{Additional Playground Visualizations}\label{app:more_viz}
\begin{figure*}[h!]
    \centering
    \begin{subfigure}[b]{0.48\textwidth}
        \centering
        {\expandafter\includegraphics\expandafter[width=\linewidth, trim=80 150 100 60, clip]{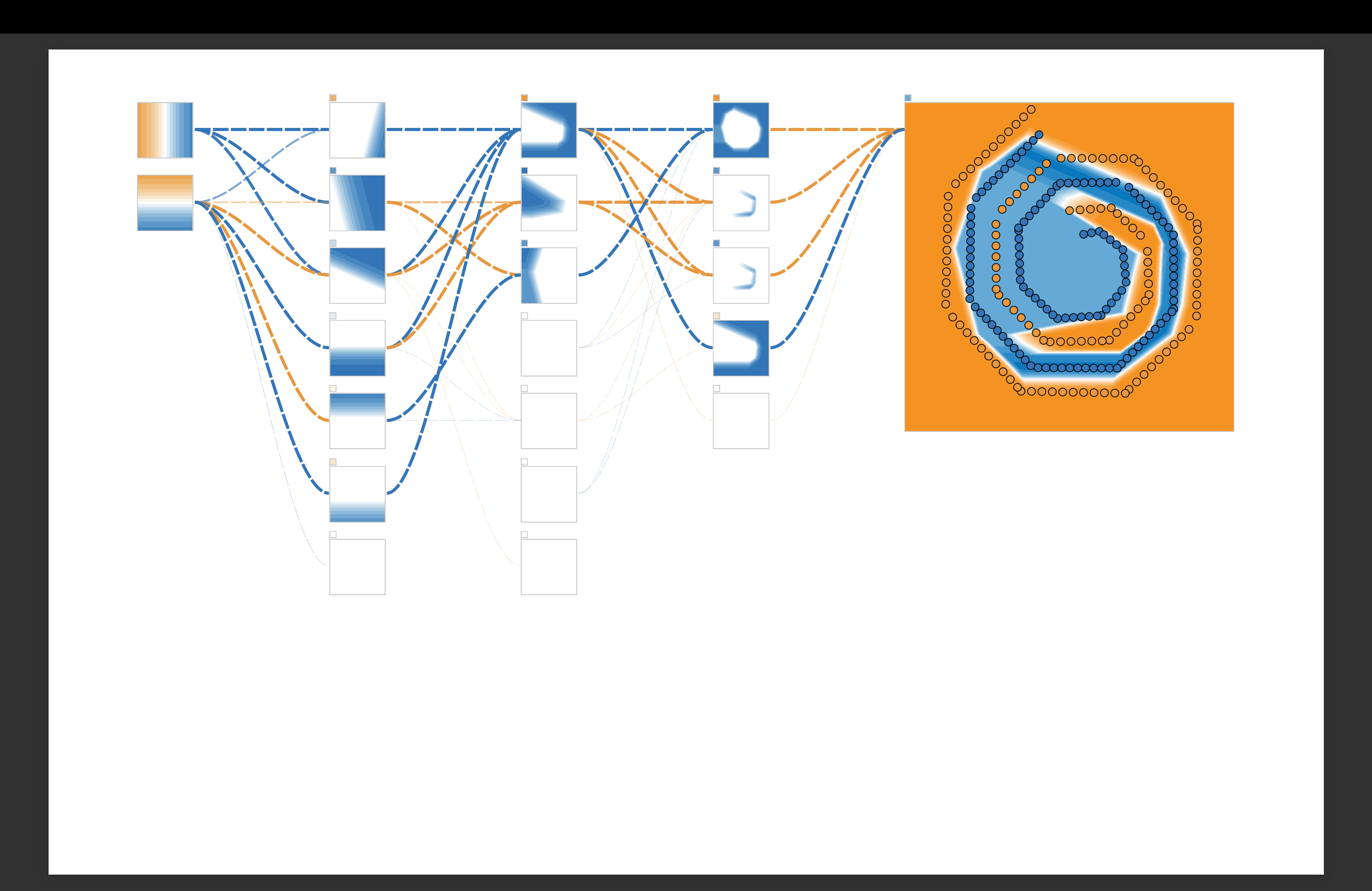}}
        \caption{Model attained using FORCE. 62 nonzeros and 90.47\% accuracy. Observe there are a significant number of small magnitude weights.}
    \end{subfigure}
    \hfill
    \begin{subfigure}[b]{0.48\textwidth}
        \centering
        {\expandafter\includegraphics\expandafter[width=\linewidth, trim=80 150 100 60, clip]{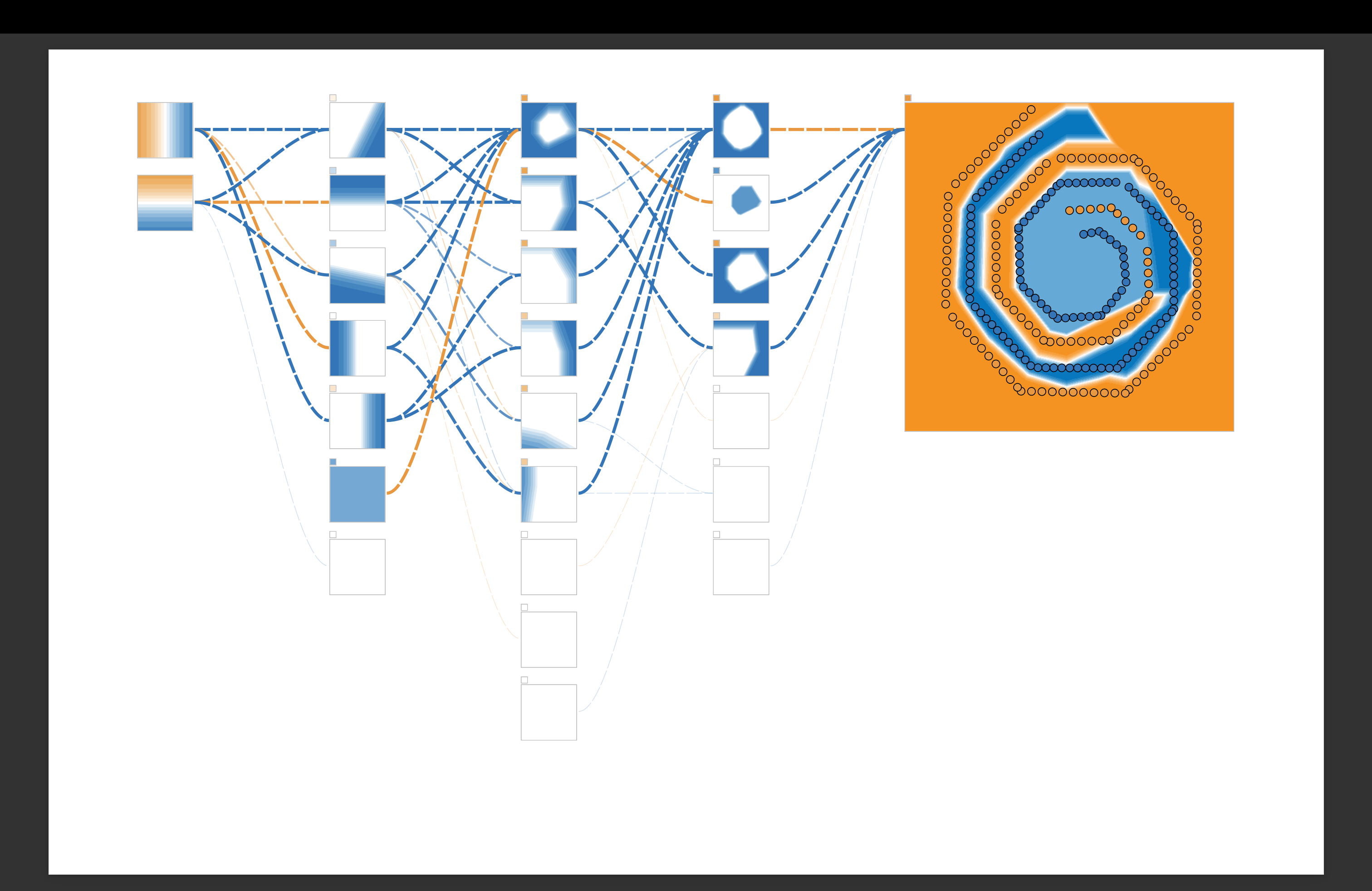}}
        \caption{Model attained using GraSP. 66 nonzeros and 94.52\% accuracy. Significant number of disconnected paths.\\}
    \end{subfigure}

    \begin{subfigure}[b]{0.48\textwidth}
        \centering
        {\expandafter\includegraphics\expandafter[width=\linewidth, trim=80 80 100 60, clip]{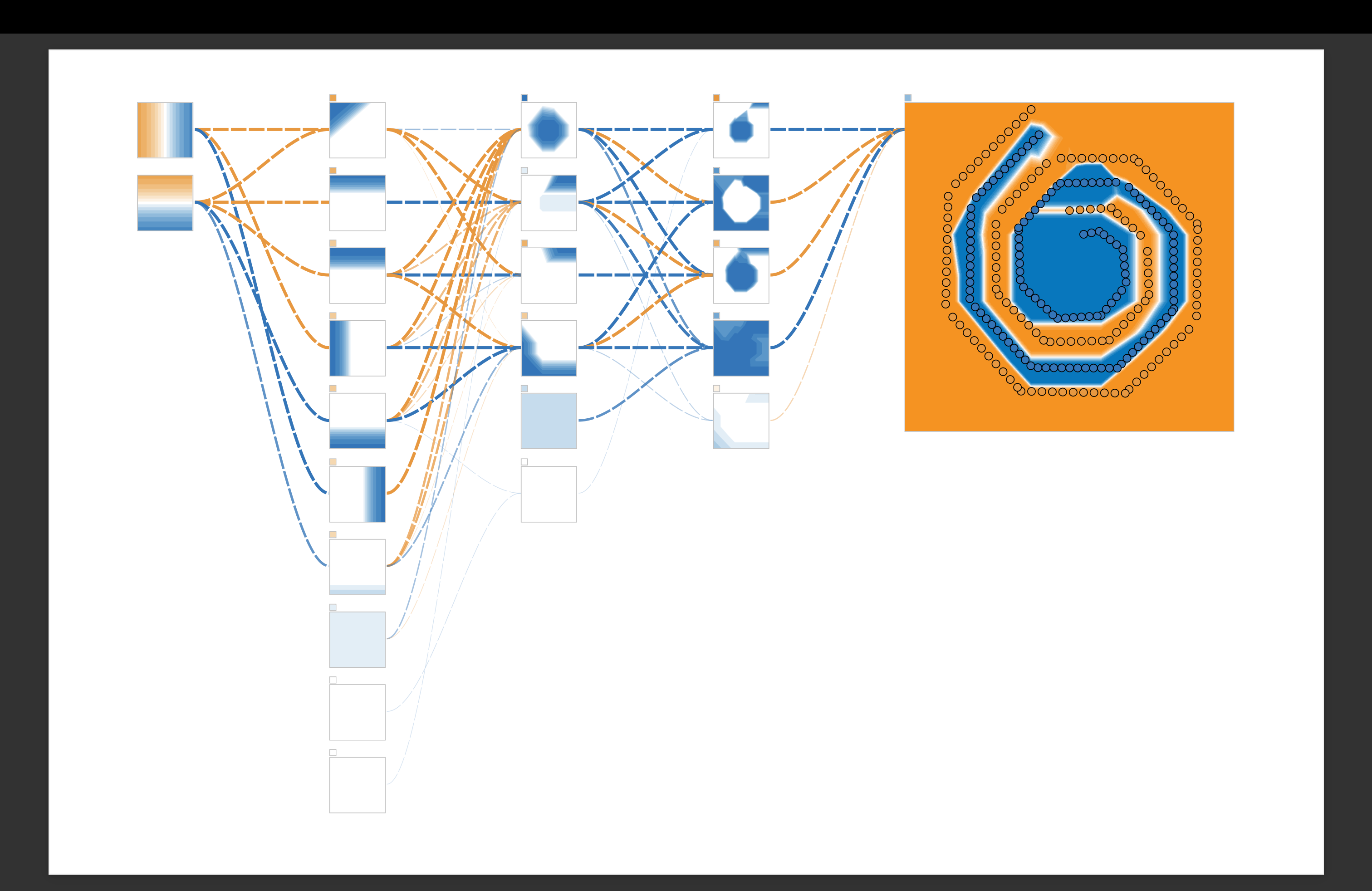}}
        \caption{Model attained using Iter SNIP. 78 nonzeros and 99.69\% accuracy. Significant number of disconnected paths.}
    \end{subfigure}
    \hfill
    \begin{subfigure}[b]{0.48\textwidth}
        \centering
        {\expandafter\includegraphics\expandafter[width=\linewidth, trim=80 80 100 60, clip]{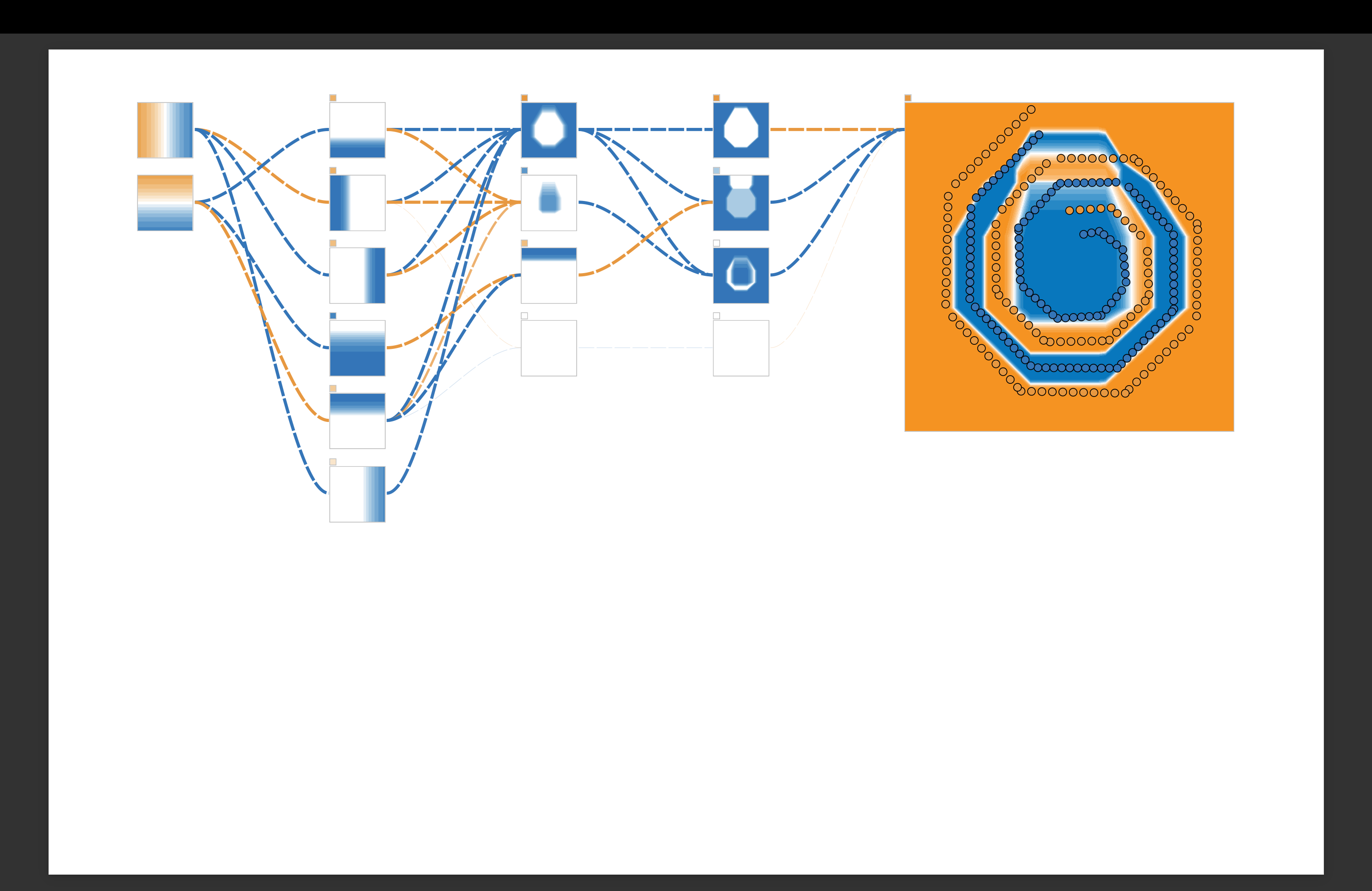}}
        \caption{Model attained using LTH. 44 nonzeros and 96.23\% accuracy.\\}
    \end{subfigure}
    \caption{Visualizations of four-layer MLPs of width 16 that were attained by FORCE, GraSP, Iter SNIP, and LTH. While all pruning algorithms suffer from disconnected paths to some degree, some do appear to be more robust than others to the issue.}
    \label{fig:prune_16_1}
\end{figure*}

\begin{figure*}[h!]
    \centering
    \begin{subfigure}[b]{0.48\textwidth}
        \centering
        {\expandafter\includegraphics\expandafter[width=\linewidth, trim=80 80 100 60, clip]{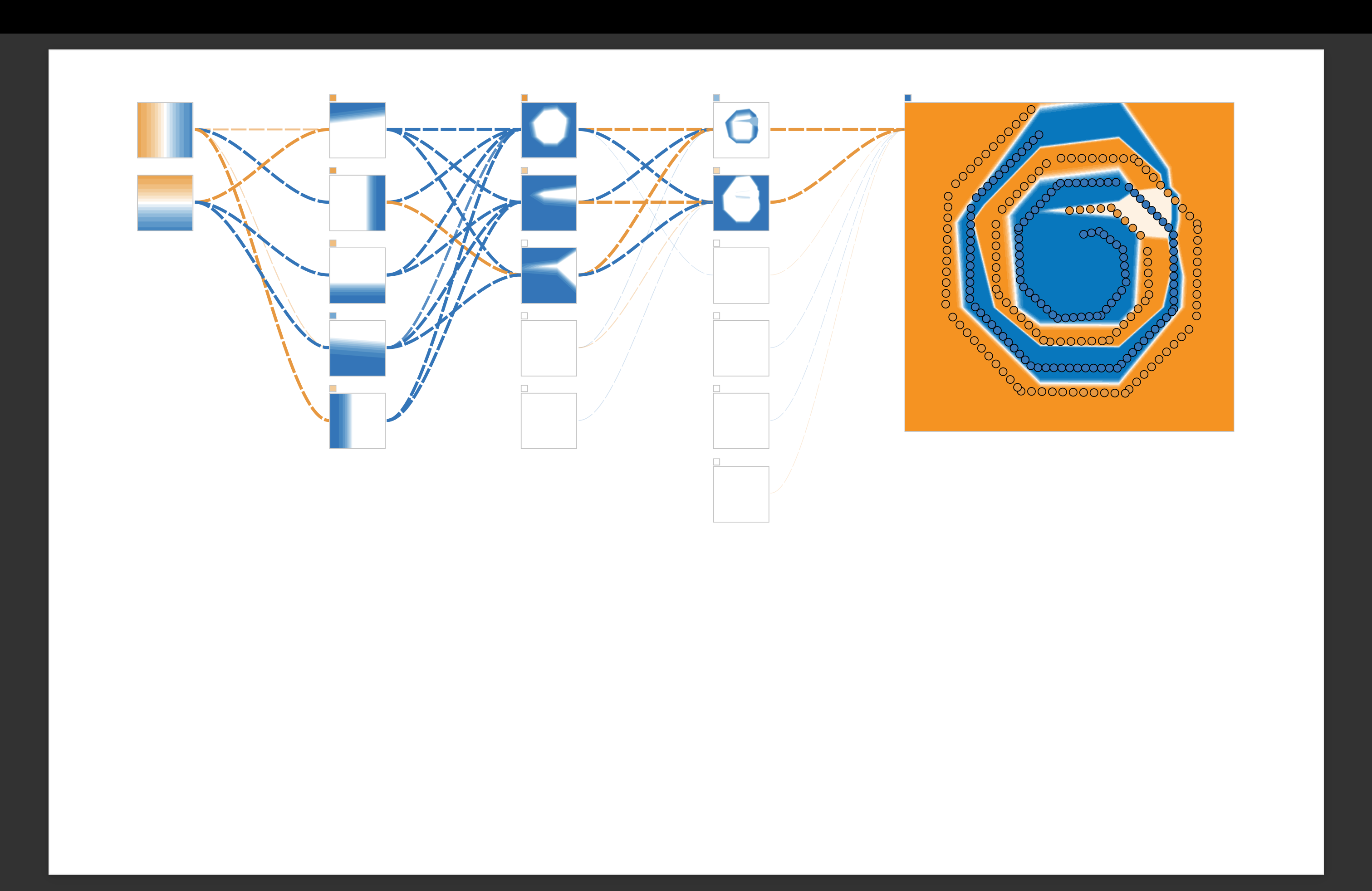}}
        \caption{Model attained using RigL. 52 nonzeros and 97.04\% accuracy. Significant number of disconnected paths.}
    \end{subfigure}
    \hfill
    \begin{subfigure}[b]{0.48\textwidth}
        \centering
        {\expandafter\includegraphics\expandafter[width=\linewidth, trim=80 80 100 60, clip]{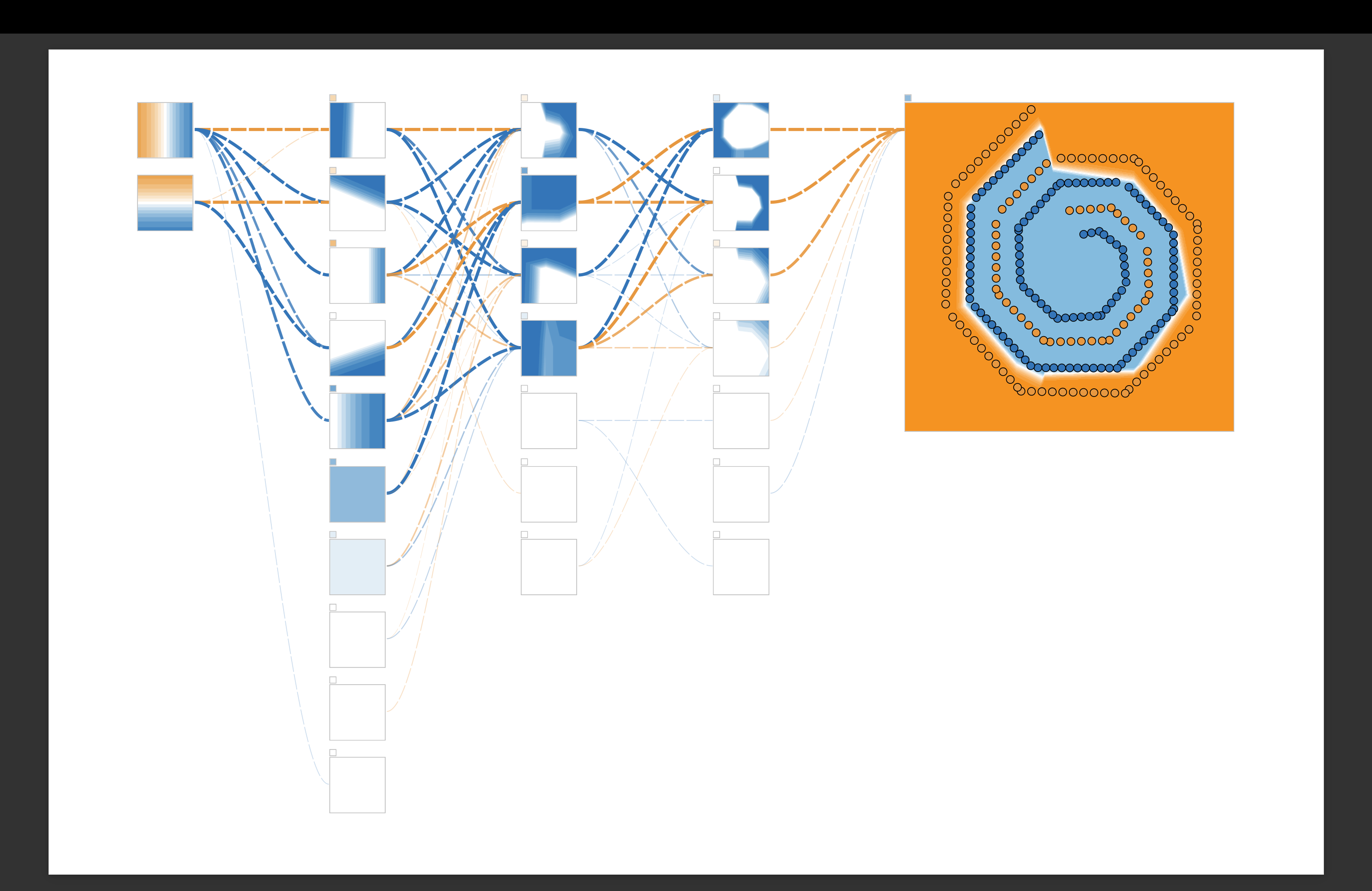}}
        \caption{Model attained using SNIP. 79 nonzeros and 81.30\% accuracy. Significant number of disconnected paths.}
    \end{subfigure}

    \begin{subfigure}[b]{0.48\textwidth}
        \centering
        {\expandafter\includegraphics\expandafter[width=\linewidth, trim=80 100 100 60, clip]{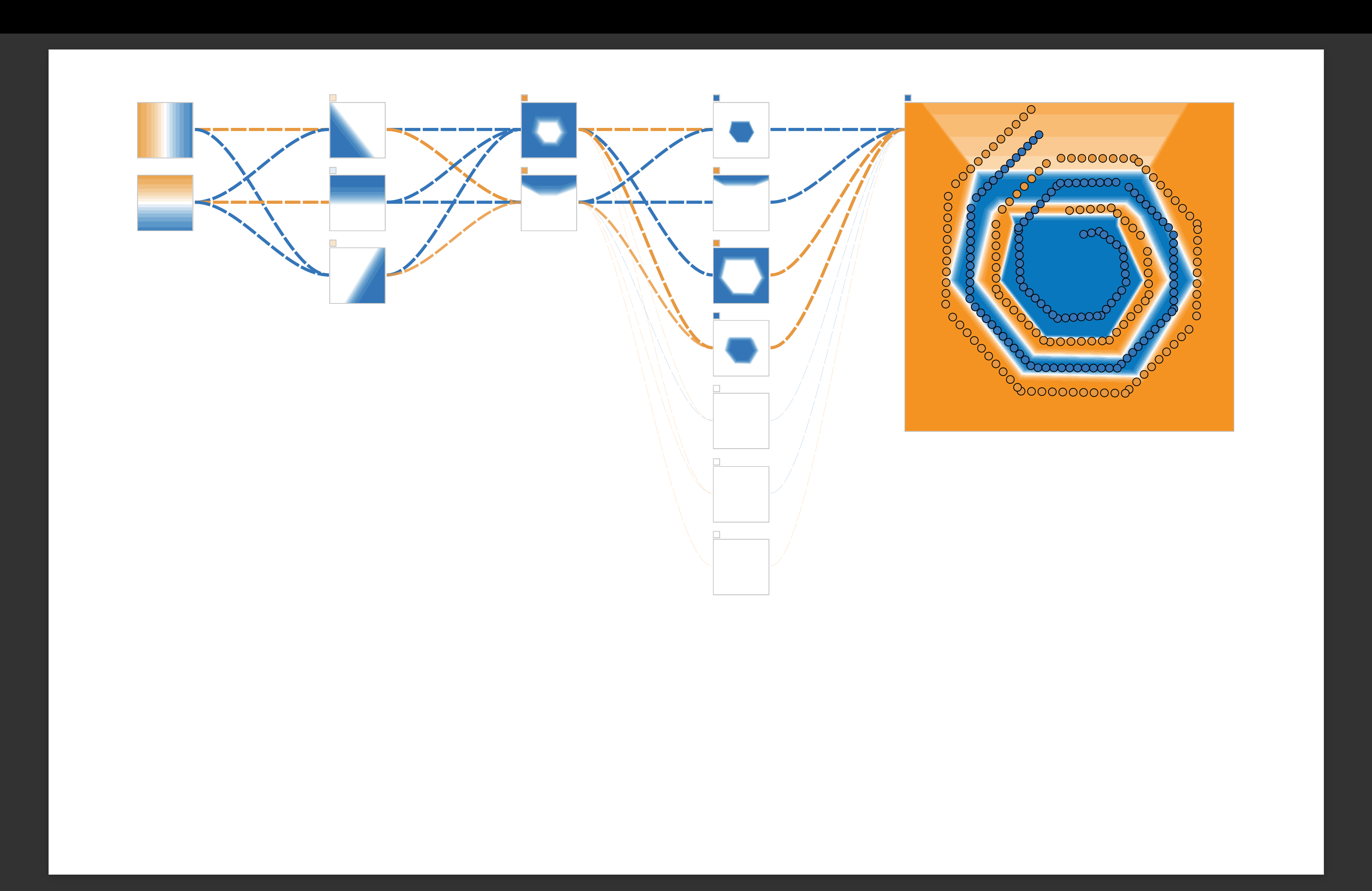}}
        \caption{Model attained using SynFlow. 42 nonzeros and 94.76\% accuracy. Presence of small magnitude weights.}
    \end{subfigure}

    \caption{Various visualizations of four-layer MLPs of width 16 that were attained by RigL, SNIP, SynFlow.}
    \label{fig:prune_16_2}
\end{figure*}

See Figures \ref{fig:prune_16_1}, \ref{fig:prune_16_2} to see more visualizations of models that were obtained from pruning a width 16, four-layer MLP that was randomly initialized. 
\newcommand{\blueitaliccomment}[1]{\textcolor{gray}{#1}}
\SetCommentSty{blueitaliccomment}

\begin{algorithm*}[h!]
\caption{Combinatorial Search}
\label{alg:combinatorial}

\SetKwFunction{Eligible}{\textsc{Eligible}}
\SetKwFunction{BinaryToDecimal}{\textsc{BinaryToDecimal}}
\SetKwFunction{EligibleMasks}{\textsc{EligibleMasks}}
\SetKwFunction{PadWithZeros}{\textsc{PadWithZeros}}
\SetKwFunction{Accuracy}{\textsc{Accuracy}}
\SetKwFunction{Stack}{\textsc{Stack}}
\SetKwFunction{MLP}{\textsc{MLP}}
\SetKwFunction{Rows}{\textsc{Rows}}
\SetKwFunction{Stack}{\textsc{Stack}}

\DontPrintSemicolon
\KwIn{Desired Accuracy, $\rho$, and maximal number of channels per layer, $D$.}
\KwOut{The set of all possible model masks $S$.}

\BlankLine

\Fn{\CombinatorialSearch{$\rho, D$}}{
    \tcp{Phase One: Loop through all possible numbers of neurons in each layer.}
    $N \leftarrow \{ \}$\;
    \tcp{Two-dimensional input and one-dimensional output.}
    $d^{[0]} = 2, d^{[4]}=1$\;
    \For{($d^{[1]}, d^{[2]}, d^{[3]}) \in \{1,\ldots,D\}^{3}$}{
        $\theta \leftarrow \MLP(\text{depth}{=}4, \text{width}{=}D)$\;
        \tcp{Mask layers according to neuron configuration.}
        \For{$\ell \in \{1,2,3,4\}$}{
        $\mW^{[\ell]}[\::\:,\: d^{[\ell]}:D] = 0, \quad \mW^{[\ell]}[d^{[\ell-1]}:D, \::\:] = 0, \quad \vb^{[\ell]}[d^{[\ell]}:D] = 0$\;
        }
        \uIf{$\Accuracy(\theta) > \rho$}{
            $N \leftarrow N \cup \{(d^{[1]}, d^{[2]}, d^{[3]})\}$
        }
    }
    \tcp{Find successful configuration that minimizes model nonzeros.}
    $d^{[1]}, d^{[2]}, d^{[3]} \leftarrow \argmin_{(d^{[1]}, d^{[2]}, d^{[3]}) \in N} (2*d^{[1]}+d^{[1]}) +( d^{[1]}*d^{[2]}+d^{[2]}) + (d^{[2]}*d^{[3]}+ d^{[3]})+ (d^{[3]}*1+1)$\;
    \tcp{Phase Two: Generate eligible masks for each layer.}
    \For{$\ell \in \{1,2,3,4\}$}{
        $\text{supp}^{[\ell]} \leftarrow \EligibleMasks(d^{[\ell-1]}, d^{[\ell]})$
    }
    \KwRet $\{(\vs^{[1]}, \vs^{[2]}, \vs^{[3]}, \vs^{[4]}) \mid  \vs^{[1]} \in \text{supp}^{[1]}, \vs^{[2]} \in \text{supp}^{[2]}, \vs^{[3]} \in \text{supp}^{[3]}, \vs^{[4]} \in \text{supp}^{[4]} \}$
}

\BlankLine

\Fn{\EligibleMasks{$d^{[in]}, d^{[out]}$}}{
    $\text{supp} \leftarrow \{ \}$\;
    \tcp{Calculate min and max possible nonzeros for each layer.}
    $\min = \max(d^{[in]}, d^{[out]})$ \;
    $\max = d^{[in]}\cdot d^{[out]}$\;
    \tcp{Loop through all nonzero counts for the layer's weights.}
    \nl\For{$n \in \{\min, \ldots, \max\}$ }{
            \tcp{Loop over all row-wise nonzero distributions.}
            \nl\For{
            $(k_1, \ldots, k_{d^{[out]}}) \in \left\{ \vk \in \{1,2,\ldots, d^{[in]}\}^{d^{[out]}} \mid \sum_{i=1}^{d^{[out]}} k_i = n \right\}$}{                
                \tcp{$\gB(k) = \left\{ v \in \{0,1\}^{d^{[in]}} \mid \sum_{i=1}^{d^{[in]}} v_i = k \right\}$}
                \tcp{Loop over all possible masks for each row in the layer.}
                \nl\For{$\vs_1 \in \gB(k_1) \land \ldots \land \vs_{d^{[out]}} \in \gB(k_{d^{[out]}})$}{
                    \nl\If{$\Eligible(\Stack(\vs_1,\ldots,\vs_{d^{[out]}}))$}{
                        $\text{supp} \leftarrow \text{supp} \cup \{\PadWithZeros(\vs_1,\ldots,\vs_{d^{[out]}})\}$\;
                    }
                }
            }
        }
    \KwRet $\text{supp}$
}

\BlankLine

\Fn{\Eligible{$\mS$}}{
    \uIf{$\mS$ contains zero columns}{
            \KwRet False
        }
    \tcp{Ensure non-increasing nonzeros across the rows.}
    \For{$i \in \{1,\ldots,\Rows(\mS)-1\}$}{
        \uIf{$\|\mS_{i, :}\|_0 > \|\mS_{i+1, :}\|_0$}{
            \KwRet False
        }
        \ElseIf{$\|\mS_{i,:}\|_0 = \|\mS_{i+1,:}\|_0$}{
            \If{$\BinaryToDecimal(\mS_{i,:}) > \BinaryToDecimal(\mS_{i+1,:})$}{
                \KwRet False
            }
        }
    }
    \KwRet True
}
\end{algorithm*}

\section{Technical Details of Sparse Model Visualization}
The visualization tool operates in two parallel threads. The first thread uses PyTorch to produce the input variables, post-activation states, and model output by inputting a $512\times 512$ grid of evenly spaced points in the square $[-2.25, 2.25] \times [-2.25, 2.25]$. It saves these tensors as well as the model parameters as global variables. The second thread runs a Flask application that visualizes these tensors using a blend of HTML and JavaScript. Specifically, the squares representing the input variables, post-activation states, and model output are illustrated using HTML Canvases for efficiency. Meanwhile, the connections denoting the weight entries are visualized with the JavaScript library D3, through Bezier curves with two control points.

\section{Increasing Width Beyond \num{16}}\label{app:width_not_needed}
In Section \ref{sec:fixed_init}, the first phase of the combinatorial search identified a structured sparsity mask using the width \num{16} MLP, comprising \num{50} nonzero parameters. Suppose we conduct a subsequent combinatorial search with an increased width of \num{17}, which uncovers a solution not attainable at width \num{16}. In this scenario, the structured sparsity mask would necessitate a minimum of \num{57} nonzeros. This requirement breaks down as follows: the first layer would need $2\times1 + 1$ parameters, the second layer $1\times 1 + 1$ parameters, the third layer $1\times 17 + 17$ parameters, and the classifier layer $17\times 1+1$ parameters, cumulatively resulting in a total of \num{57} parameters.

\section{Detailed Description of Combinatorial Search Algorithm}
The first loop in the function $\EligibleMasks(d_{in}, d_{out})$ (annotated by line 1) evaluates every conceivable quantity of nonzero elements in the weight matrix symbolized by $n$. Since each row and column contains at least one nonzero element, the minimal count of nonzero elements in the layer is determined by $\min = \max(d_{in}, d_{out})$. It's also self-evident that this count cannot exceed the product $\max = d_{in}\cdot d_{out}$.

The second loop (annotated by line 2) explores all eligible counts of nonzero elements in each row of the weight, labeled as $k_i$. This exploration ensures that the sum of nonzero elements across different rows, $k_i$, equates to the total number of nonzero elements in the entire weight matrix, $n$.

The third and concluding nested loop (annotated by line 3) peruses all potential masks of size $k_i$ for each row in the weight matrix, across all rows.

The innermost if statement (annotated by line 4) selects from all row permutations the specific arrangement that results in a sequentially decreasing count of nonzero elements. If two rows have an equal count of nonzeros, the algorithm converts the binary vector representations of these masks into their decimal equivalents and arranges them in descending order based on these decimal values. Finally, it does the final check to verify that the mask is devoid of zero columns and if deemed as an eligible mask, the mask is padded with zeros so that its dimensions match the weight matrix.

\clearpage 
\section{Overparameterization Leads to Disconnected Paths}\label{app:overparam_proof}
\begin{theorem}
    Consider an $L$-layer multilayer perceptron with weights $\mW^{[1]}, ..., \mW^{[L]}$ where $\mW^{[1]}\in \mathbb{R}^{d\times w}, \mW^{[2]}, ..., \mW^{[L-1]}\in \mathbb{R}^{w \times w}, \mW^{[L]} \in \mathbb{R}^{w \times C}$. Suppose the model is randomly pruned such that $n_{\ell}$ nonzero weights remain in $\mW^{[\ell]}$. Assume that $L \geq 4$ and \(w > \max_{\ell\in \{2,..., L-2\}}( n_{\ell} + n_{\ell + 1})\). Then the probability that there is no connected path in the model tends to one when the width of the model goes to infinity. 
\end{theorem}
\textbf{Proof:} For the proof, we will use the notation that $\mW^{[\ell]}$ denotes the unpruned weight matrix, $\mM^{[\ell]}$ denotes the mask generated by randomly pruning layer $\mW^{[\ell]}$, and $\tilde \mW^{[\ell]} = \mM^{[\ell]}\odot \mW^{[\ell]}$ denotes the pruned weight matrix. It is clear that if there are no connected paths between the two layers $\tilde \mW^{[\ell]}$ and $\tilde \mW^{[\ell+1]}$, there are no connected paths in the whole pruned network. 

\noindent
Let $A^{[\ell]}_k$ denote the set of masks satisfying:
    $$A^{[\ell]}_k = \{\mM^{[\ell]} \in \mathbb{R}^{w\times w} \; : \; \mM^{[\ell]}_{1, \cdot } \neq 0, ..., \mM^{[\ell]}_{k, \cdot } \neq 0, \mM^{[\ell]}_{k+1, \cdot} =0 , ..., \mM^{[\ell]}_{w, \cdot } =0\} $$
For $2 \leq \ell \leq L-2$, we can express the probability that there are no connected paths between $\tilde \mW^{[\ell]}$ and $\tilde \mW^{[\ell+1]}$ as follows:
    \begin{align*}
    \sum_{k=1}^{n_{l}} \binom{w}{k} \mathbb{P}(\text{No Connected Paths between } \tilde \mW^{[\ell]} \text{ and } \tilde \mW^{[\ell+1]} | \mM^{[\ell]} \in A^{[\ell]}_k) \cdot \mathbb{P}(\mM^{[\ell]} \in A^{[\ell]}_k)
    \end{align*}
For there to be no connected paths between $\tilde \mW^{[\ell]}$ and $\tilde \mW^{[\ell+1]}$, that means that all $n_{\ell+1}$ nonzero entries in $\mM^{[\ell+1]}$ must lie in the remaining $w-k$ columns that are not aligned with the $k$ nonzero rows of $\mM^{[\ell]}$. Thus, we find that
$$\mathbb{P}(\text{No Connected Paths between } \tilde \mW^{[\ell]} \text{ and } \tilde \mW^{[\ell+1]} | \mM^{[\ell]} \in A^{[\ell]}_k) = \sum_{r=1}^{n_{\ell+1}}\frac{\binom{w-k}{r} \cdot |A^{[\ell+1]}_{r}|}{\binom{w^2}{n_{\ell+1}}} $$
Combining everything together, we get that 
$$\mathbb{P}(\text{No Connected Paths between } \tilde \mW^{[\ell]} \text{ and } \tilde \mW^{[\ell+1]}) = \sum_{k=1}^{n_{\ell}}\left[\frac{\binom{w}{k} \cdot |A^{[\ell]}_k|}{\binom{w^2}{n_{\ell}}} \cdot \sum_{r=1}^{n_{\ell+1}}\frac{\binom{w-k}{r}\cdot |A^{[\ell+1]}_{r}|}{\binom{w^2}{n_{\ell+1}}} \right] $$
To see that this term is tending to one as $w\to \infty$, notice that 
$$ \sum_{k=1}^{n_{\ell}}\binom{w}{k}\cdot |A_{k}^{[\ell]}| = \binom{w^2}{n_{\ell}}$$
Utilizing this, we can bound the probability as follows:
\begin{align*}
    \sum_{k=1}^{n_{\ell}}\left[\frac{\binom{w}{k} \cdot |A^{[\ell]}_k|}{\binom{w^2}{n_{\ell}}} \cdot \sum_{r=1}^{n_{\ell+1}}\frac{\binom{w-k}{r}\cdot |A^{[\ell+1]}_{r}|}{\binom{w^2}{n_{\ell+1}}} \right] &= \sum_{k=1}^{n_{\ell}}\left[\frac{\binom{w}{k} \cdot |A^{[\ell]}_k|}{\binom{w^2}{n_{\ell}}} \cdot \sum_{r=1}^{n_{\ell+1}}\frac{\binom{w}{r}\cdot \prod_{j=0}^{k-1} \left(\frac{w-j-r}{w-j}\right) \cdot |A^{[\ell+1]}_{r}|}{\binom{w^2}{n_{\ell+1}}} \right] \\
    &\geq \left(\frac{w-n_{\ell} + 1 - n_{\ell+1}}{w}\right)^{n_{\ell}-1} 
 \sum_{k=1}^{n_{\ell}}\left[\frac{\binom{w}{k} \cdot |A^{[\ell]}_k|}{\binom{w^2}{n_{\ell}}} \cdot \sum_{r=1}^{n_{\ell+1}}\frac{\binom{w}{r}\cdot |A^{[\ell+1]}_{r}|}{\binom{w^2}{n_{\ell+1}}} \right] \\
 &=  \left(\frac{w-n_{\ell} + 1 - n_{\ell+1}}{w}\right)^{n_{\ell}-1} 
\end{align*}
arriving at the following set of inequalities:
\begin{align*}
    &\left(\frac{w-n_{\ell} + 1 - n_{\ell+1}}{w}\right)^{n_{\ell}-1} \leq \mathbb{P}(\text{No Connected Paths between } \tilde \mW^{[\ell]} \text{ and } \tilde \mW^{[\ell+1]}) \leq 1
\end{align*}
Applying squeeze theorem, we get that the probability is tending to one as $w\to \infty$. 
\clearpage
\newpage

\begin{figure*}
    \centering
    \PlotFigure{acc_vs_nnz_16}{Model NNZ}{Test Acc}{true}{0.8}
    \caption{Scatter plots containing the best performing run for each pruning approach at a given number of nonzeros in the model. Pareto frontiers depicted in Figure \ref{fig:acc_vs_nnz_16} are interpolated from the datapoints depicted in this plot.}
    \label{fig:scatter_acc_vs_nnz_16}
\end{figure*}

\begin{figure*}
    \centering
      \begin{subfigure}[b]{0.45\linewidth}
          \centering
          \PlotFigure{recovery_995/recovery_6}{Model NNZ1}{Test Acc1}{false}{0.95}
          \caption{Models Given Optimal Width of 6}
      \end{subfigure}
      \begin{subfigure}[b]{0.45\linewidth}
          \centering
          \PlotFigure{recovery_995/recovery_16}{Model NNZ1}{Test Acc1}{true}{0.95}
          \caption{Width 16}
      \end{subfigure}
    \caption{Scatter plots containing the best performing run for each pruning approach at a given number of nonzeros in the model. Pareto frontiers depicted in Figure \ref{fig:recovery} are interpolated from the datapoints depicted in this plot.}
    \label{fig:scatter_recover_995}
\end{figure*}

\begin{figure*}
    \centering
      \begin{subfigure}[b]{0.45\linewidth}
          \centering
          \PlotFigure{recovery_95/recovery_3}{LastNNZ}{LastTest}{false}{0.95}
          \caption{Models Given Optimal Width of 3}
      \end{subfigure}
      \begin{subfigure}[b]{0.45\linewidth}
          \centering
          \PlotFigure{recovery_95/recovery_16}{LastNNZ}{LastTest}{true}{0.95}
          \caption{Width 16}
      \end{subfigure}
    \caption{Scatter plots containing the best performing run for each pruning approach at a given number of nonzeros in the model. Pareto frontiers depicted in Figure \ref{fig:recovery_95} are interpolated from the datapoints depicted in this plot.}
    \label{fig:scatter_recover_95}
\end{figure*}

\end{document}